\documentclass[oneside, sn-mathphys ]{sn-jnl}
\usepackage[authoryear]{natbib}
\bibpunct{(}{)}{,}{a}{}{;}

\usepackage{amsmath} 
\usepackage{graphicx}
\usepackage{multirow}%
\usepackage{amsmath,amssymb,amsfonts}%
\usepackage{amsthm}%
\usepackage{mathrsfs}%
\usepackage[title]{appendix}%
\usepackage{xcolor}%
\usepackage{textcomp}%
\usepackage{manyfoot}%
\usepackage{booktabs}%
\usepackage{algorithm}%
\usepackage{algorithmicx}%
\usepackage{algpseudocode}%
\usepackage{listings}%
\usepackage{lineno} 
\usepackage{booktabs}
\usepackage[normalem]{ulem}
\useunder{\uline}{\ul}{}
\usepackage{tabularx,ragged2e}
\usepackage{caption}
\usepackage{tikz}
\usepackage[utf8]{inputenc}
\usepackage{geometry}
\usepackage{titlesec}
\usepackage{graphicx}
\usepackage{graphicx} 

\usepackage{adjustbox}
\usepackage{microtype}  
\usepackage{url} 

\raggedright
\title[Article Title]{A Survey of the Self Supervised Learning Mechanisms for Vision Transformers}

\author[1,2*]{\fnm{Asifullah} \sur{khan}}\email{asif@pieas.edu.pk}

\author[3]{\fnm{Anabia} \sur{Sohail}}
\author[4]{\fnm{Mustansar} \sur{Fiaz} 
}
\author[5]{\fnm{Mehdi} \sur{Hassan}}
\author[6]{\fnm{Tariq Habib} \sur{Afridi}
}

\author[1]{\fnm{Sibghat Ullah} \sur{Marwat}}
\author[7]{\fnm{Farzeen} \sur{Munir}}
\author[1]{\fnm{Safdar Ali}}
\author[8]{ \fnm{Hannan} \sur{Naseem}}
\author[9]{ \fnm{Muhammad Zaigham} \sur{Zaheer}}
\author[10]{\fnm{Kamran} \sur{Ali}}
\author[6,11]{\fnm{Tangina} \sur{Sultana}}
\author[12]{\fnm{Ziaurrehman} \sur{Tanoli}}
\author[1]{\fnm{Naeem} \sur{Akhter}
}

\affil[1]{\orgaddress{Pattern Recognition Lab, DCIS, PIEAS, Nilore, Islamabad, 45650, Pakistan.}}
\affil[2]{ \orgaddress{PIEAS Artificial Intelligence Center (PAIC), PIEAS, Nilore, Islamabad, 45650, Pakistan.}}
\affil[3]{\orgname{Center of Secure Cyber-Physical Security Systems, Khalifa University, Abu Dhabi, United Arab Emirates}
\orgaddress{\country{UAE}}}

\affil[4]{\orgname{IBM Research}}

\affil[5]{\orgdiv{Department of Computer Science} \orgname{Air University} \city{Islamabad} \country{Pakistan} 
}
\affil[6]{\orgdiv{Department of Computer Science and Engineering}, \orgname{Kyung Hee University (Global Campus)}, \orgaddress{\street{1732 }, \city{Yongin}, \postcode{17104}, \state{Gyeonggi-do} \country{Republic of Korea}}}

\affil[7]{\orgname{Department of Electrical Engineering and Automation, Aalto University Finland \& Finnish Center of Artificial Center, \country{Finland}}}
\affil[8]{\orgdiv{Faculty of Engineering and Green Technology}, \orgname{Universiti Tunku Abdul Rahman}, \orgaddress{\street{}\city{}\postcode{}\state{}\country{Malaysia}}}

\affil[9]{\orgdiv{Computer Vision Department}, \orgname{Mohamed Bin Zayed University of Artificial Intelligence},\orgaddress{\country{UAE}}}

\affil[10]{\orgname{Foundation for Advancement of Science and Technology (FAST)},\orgaddress{\city{Karachi}}}

\affil[11]{\orgdiv{Department of Electronics and Communication Engineering}, \orgname{Hajee Mohammad Danesh Science and Technology University}, \orgaddress{\street{}\city{}\postcode{}\state{}\country{Bangladesh}}}

\affil[12]{ Institute for Molecular Medicine Finland (FIMM), HiLIFE, University of Helsinki, Finland.}

\begin{document}

\clearpage
\abstract{
\justifying
Advances in deep learning are re-defining how visual data is processed and understand by the machines. 
Vision Transformers (ViTs) have recently demonstrated prominent performance in computer vision related tasks. However, their performance improves with increasing numbers of labeled data, indicating reliance on labeled data. Humanly annotated data are difficult to acquire and thus shifted the focus from traditional annotations to unsupervised learning strategies that learn structures inside the data. In response to this challenge, self-supervised learning (SSL) has emerged as a promising technique. SSL utilize inherent relationships within the data as a form of supervision. This technique can reduce the dependence on manual annotations and offers a more scalable and resource-effective approach to training models. Taking these strengths into account, it is necessary to assess the combination of SSL methods with ViTs, especially in the cases of limited labeled data. Inspired by this evolving trend, this survey aims to systematically review SSL mechanisms tailored for ViTs. We propose a comprehensive taxonomy to classify SSL techniques based on their representations and pre-training tasks. Furthermore, we highlighted the motivations behind the study of SSL, reviewed prominent pre-training tasks, and highlight advancements and challenges in this field. Furthermore, we conduct a comparative analysis of various SSL methods designed for ViTs, evaluating their strengths, limitations, and applicability to different scenarios.
}




\maketitle
\section{Introduction}

\justifying


Deep learning-based algorithms have achieved remarkable success across various disciplines, particularly in computer vision \citep{kalluri2025computer,ren2024intelligent, zhang2022computerVision, mendes2025you} and natural language processing \citep{bosma2025dragon, min2023recentNLP}. These models often utilize a pretraining task on large datasets to improve performance. This initial step typically serves as an initialization point, after which the model is fine-tuned for specific downstream tasks. Among pre-training strategies for deep learning, self-supervised learning (SSL) has emerged as a prominent approach. SSL approach is driven by two key motivations. Firstly, network trained on extensive datasets learn transferable patterns that reduce overfitting during training. Secondly, parameters derived from large-scale datasets provide effective initialization, allowing faster convergence between different applications \citep{rani2023self}.
%
Despite the abundance of unlabeled web data in the era of big data, developing high-quality labeled datasets through human annotation remains expensive.
For instance, platforms such as Scale.com \citep{scale.com} charge approximately \$6.4 per image for the labeling task such as image segmentation. Building an image segmentation dataset containing over a million high-quality samples, like JFT-300M, can cost millions of dollars. Usually, this process is time-intensive and inefficient \citep{gui2024survey}. Moreover, methods of supervised learning can easily learn misleading connections within dataset, which can result in mistakes and thus may be more sensitive to adversarial exploitations.
To address these challenges, strategies like active learning \citep{activelearning}, semi-supervised learning \citep{semiSupervised}, and SSL \citep{ssl} have gained traction. Recently, transformers \citep{vaswani2017attention} have emerged as an effective neural network (NN) architecture to leverage deeper insights into input data; specifically, ViTs \citep{dosovitskiy2020image} are considered the alternative to CNNs \citep{lecun1998gradient, alipour2025leveraging, dai2025pathologyvlm, al2025computer, gonzalez2025semi, hosny2025explainable} ViTs have been successfully employed in tasks related to computer vision, including object detection \citep{li2022exploring}, recognition \citep{ranzato2021advances} and semantic segmentation \citep{thisanke2023semantic}. In particular, ViTs have demonstrated superior performance in image classification, when trained with large-volume datasets such as JFT-300M \citep{sun2017revisiting}.
In essence, SSL methods enable Vision Transformers (ViTs) to understand the underlying structures of data without supervision through pretext tasks, the self-designed objectives in a pre-training phase that encourage the model to learn significant patterns within the data itself. The motivation for applying SSL in ViTs is two-fold. Firstly, SSL enables ViTs to utilize large amounts of unlabeled data,  fostering robust and generalizable feature representations. This approach is particularly advantageous when labeled data is limited or costly to produce. Secondly, SSL facilitates pre-training ViTs on extensive datasets, which can then be fine-tuned for specific downstream tasks. This process not only improves the performance of the model but also accelerates training convergence \citep{gui2024survey}.

Several interesting surveys related to SSL have been reported. However, these surveys on SSL, serve to particular domains like recommendation systems \citep{SSLforRecommenderSystem}, graphs \citep{GraphSSL}, \citep{GraphSSL2}, sequential transfer learning \citep{pre-training}, videos \citep{SSLForVideos},  algorithms \citep{gui2024survey}, non-sequential tabular data \citep{wang2025survey} and recent advancements in using deep learning and image-based models for plant disease detection \citep{upadhyay2025deep}. A recent survey summarize the deep learning models deployed on edge devices, particularly in healthcare and computer vision \citep{xu2025edge}, highlighting the trade-offs between performance and resources usage. 
Vision Transformers (ViTs) are a prominent area of research in computer vision \citep{wang2025vision} and have recently demonstrated several breakthroughs. However, the training of ViTs remains a challenge. 

In contrast to existing surveys, this paper presents a comprihensive examination of different SSL methods proposed for ViTs, offering a structured taxonomy for pre-text techniques, a comparitive analysis with transfer learning paradigms, an overview of regularization strategies and evaluation metrics, and a discussion on prevailing challenges along with prospective research directions.

\begin{figure}[ht]
    \centering
    \includegraphics[width=13cm ,height=6.5cm]{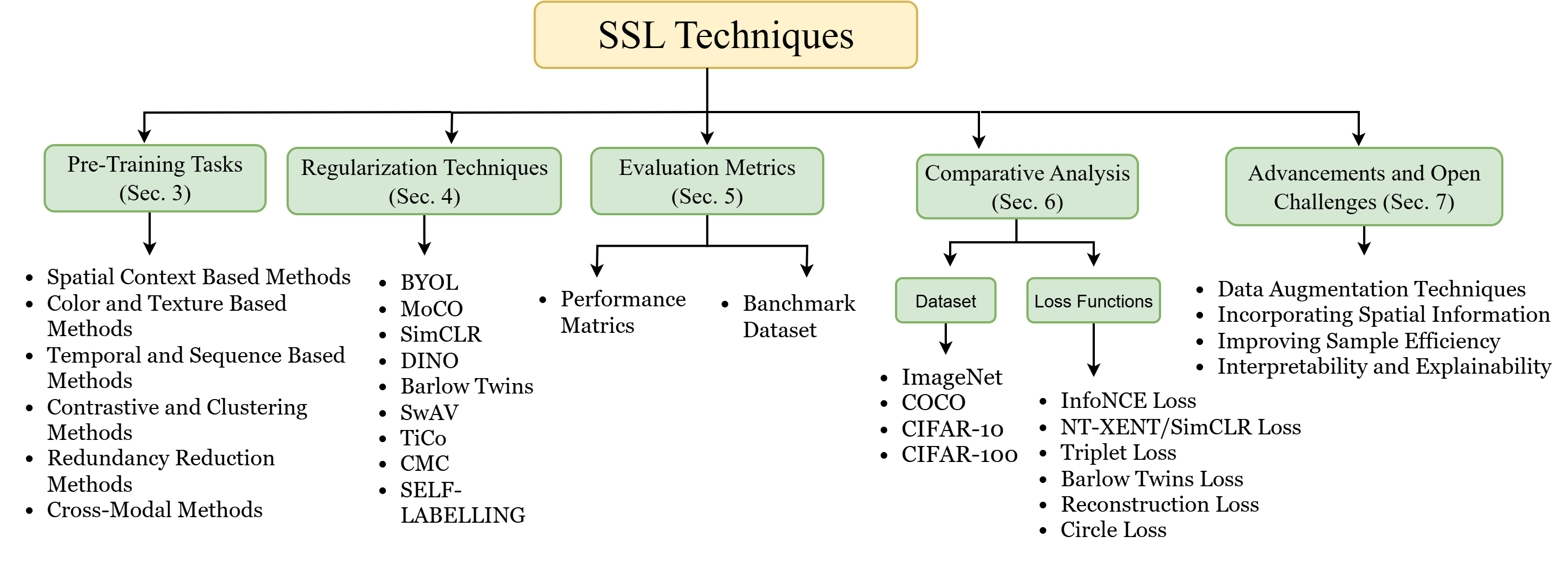}
    \vspace{\baselineskip}
    \caption{Hierarchical overview of the paper’s structure on Self-Supervised Learning (SSL) techniques.  The root node of SSL Techniques is divided further into five major sections as shown in the figure, namely: Section 3 Pre-Training Tasks (covers spatial context-based, color and texture based, temporal and sequence based, contrastive and clustering, distillation and momentum-based, redundancy reduction, and cross-modal techniques); Section 4 Regularization Techniques With representative methods being BYOL, MoCo, SimCLR, DINO, Barlow Twins, SwAV, TiCo, CMC, and self-labelling approaches being surveyed). Section 5 explains the evaluation Metrics along with benchmarks and loss functions. Section 6 discusses comparitive analysis of SSL mechanisms. Section 7 discuss about advancements and open challenges while Section 8 discusses future directions.} 
    \label{fig:intro}
\end{figure}

\newpage 


\subsection{Survey Structure}
The paper begins with an introduction to SSL and a detailed discussion on importance of SSL in ViTs. Then we discussed the recent advancements and applications of SSL across different computer vision tasks. It also presents a taxonomy for SSL, which classifies them according to how feature learning is applied within their architecture. According to this taxonomy, SSLs are divided into five groups, each representing unique way to make use of input features. Frequently used abbreviations are listed in Table. \ref{tab:abbreviations}. The structure of paper is depicted in Fig. \ref{fig:intro}. Section 1 discuss a systematic understanding of the SSL architecture, highlighting its need in ViTs and outlined the advent of SSL architectures. Proceed to section 2 which covers the advancements in SSL variants, while section 3 provides a taxonomy of the recent SSL architectures, respectively. Section 4 focuses on the regularization techniques used in SSL, particularly in the area of computer vision. Section 5 presents various metrics and benchmarks to evaluate the effectiveness of SSL, while Section 6 provides a comparative analysis of state-of-the-art SSL mechanisms that employ Vision Transformers. Section 7 highlights some open challenges in the domain of SSL, and Section 8 discusses different applications and future directions for SSL in Vision Transformers, including current challenges and potential developments. Finally, Section 9 concludes this survey.

\begin{table*}[ht]
    \centering
    \caption{Abbreviations used throughout the survey for models and techniques in self-supervised learning (SSL)}
    \label{tab:abbreviations}
    \renewcommand{\arraystretch}{1.3} 
    \setlength{\tabcolsep}{10pt} 
    \begin{tabular}{l p{9cm}} 
         \hline 
         \textbf{Abbreviation} & \textbf{Definition} \\ \hline 
          SSL   &  Self-Supervised Learning \\ 
          ViT   &  Vision Transformer \\ 
          CV    &  Computer Vision \\ 
          NLP   &  Natural Language Processing \\ 
          CNN   &  Convolutional Neural Network \\ 
          NN    &  Neural Network \\ 
          GAN   &  Generative Adversarial Network \\ 
          VAE   &  Variational Autoencoder \\ 
          BERT  &  Bidirectional Encoder Representations from Transformers \\ 
          SimCLR & Simple Framework for Contrastive Learning of Visual Representations  \\ 
          SRCL  & Semantically Relevant Contrastive Learning \\ 
          MoCo  &  Momentum Contrast \\ 
          SwAV  & Swapping Assignments between Views \\ 
          BYOL  &  Bootstrap Your Own Latent \\ 
          DeiT  & Data-efficient Image Transformer  \\ 
          CL    & Contrastive Learning \\ 
          CTransPath & Contrastive Transformer for Pathology \\ 
          MCVT  & Multi-level Contrastive Learning for Vision Transformers \\ 
          CLIP  & Contrastive Language-Image Pre-training \\ 
          DINO  & Self-Distillation with No Labels \\ 
          EsViT & Efficient Self-Supervised Vision Transformer \\ 
          FLSL  & Feature-Level Self-Supervised Learning \\ 
          COCO  & Common Objects in Context (Microsoft COCO dataset) \\ 
          R-CNN & Region-based Convolutional Neural Network \\ 
          GPT   & Generative Pre-trained Transformer \\ 
          CIFAR-10 & Canadian Institute for Advanced Research 10-class dataset \\ 
          MAE   & Masked Autoencoder \\ 
          BEiT  & Bidirectional Encoder Representation from Images \\ 
          IR    & Intermediate Representation \\ 
          EMA   & Exponential Moving Average \\ 
          GCMAE & Global Contrast Masked Autoencoder \\ 
          HSL   & Hybrid Self-Supervised Learning Framework \\ 
          HSI   & Hyperspectral Image \\ 
          CMC   & Contrastive Multiview Coding \\ 
          RPC   & Relative Predictive Coding \\ 
          CPC   & Contrastive Predictive Coding \\ 
          NAT   & Noise As Targets \\ 
          SimSiam & Simple Siamese Representation Learning \\ \hline
    \end{tabular}
\end{table*}

\clearpage

\section{Evolution of SSL}\label{secBackground}

In computer vision, SSL methods are generally segmented into contrastive, generative, and predictive approaches. Contrastive methods, such as MoCo \citep{he2020momentum} and SimCLR \citep{chen2020simple}, aim to learn patterns by contrasting positive and negative samples. Generative methods, such as GANs \citep{goodfellow2020generative} and VAEs \citep{kingma2019introduction}, focus on learning representations by generating samples from the learned representation space. Predictive methods, such as BERT \citep{devlin2018bert} and T5 \citep{raffel2020exploring}, learn representations by predicting some parts of the input data.

One SSL category is instance discrimination, in which the motto of learning is basically focused on learning patterns by differentiating each sample image from others. This method is rather challenging for large volume of datasets. Examples are SimCLR, SwAV, BYOL, MoCO \citep{chen2020simple,caron2020unsupervised, grill2020bootstrap,he2020momentum}.
\subsection{Evolution of ViTs}

In the domain of computer vision, ViTs is newly arising development motivated by the success of transformers architecture which is presented in seminal paper "Attention Is All You Need" \citep{vaswani2017attention}. Transformers has revolutionized the learning process by enabling models to learn patterns from substantial volume of unlabeled data. This approach enables global context modeling, distinguishing ViTs from traditional CNNs. Early ViT models demonstrated strong performance in image classification but faced challenges in computational efficiency and data requirements \citep{elharrouss2025vits}.

In computer vision, the traditional approach in image processing tasks is the utilization of convolutional neural networks (CNNs). However, the CNNs have limitations in capturing dependencies across long spatial distances and global context, both of which are crucial for some vision tasks. ViTs address these limitations via self-attention mechanisms to learn and understand the global context and also dependencies over long spatial distances in the input data. ViTs have also shown strong results in image classification, object detection, medical imaging, and image restoration, often surpassing CNNs when trained on large datasets \citep{haruna2025exploring}.

The initial vision transformer architecture ViT was presented in the paper "An Image Is Worth 16x16 Words" \citep{dosovitskiy2020image}. Since then, various variants of ViTs are proposed, including DeiT \citep{touvron2021training}, Swin Transformer \citep{liu2021swin}, and TNT \citep{han2021transformer}. These architectures have obtained state-of-the-art performance in different computer vision based tasks, including image classification, object segmentation and object detection.

\subsection{Importance of SSL in ViTs}\label{subsecBackSSLinVT}

SSL is especially crucial in the context of ViTs, as it empowers models to learn and understand feature representations from extensive unlabeled data. ViTs rely on extensive datasets to learn meaningful and effective representations while SSL offers a solution by enabling pretraining on large-scale datasets, allowing ViTs to acquire robust and transferable feature representations for downstream tasks \citep{ye2024cluster,yun2022patch,wang2022self,nguyen2024exploring}.

\begin{figure}[H]
    \centering
    \includegraphics[width=1\linewidth]{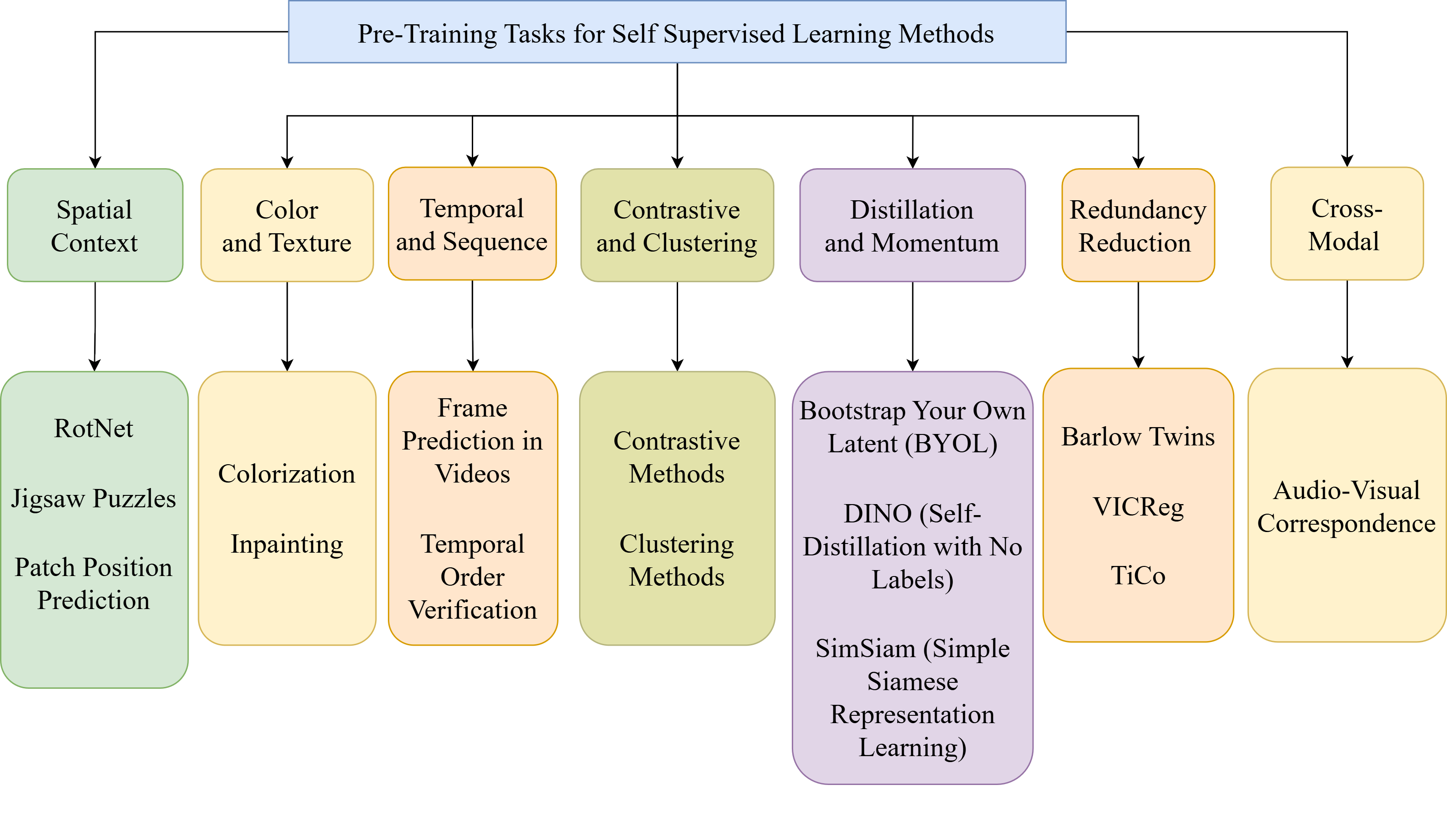}
    \caption{Pre-training tasked based categorization of SSL methods. These categorizes tries to provide a brief understanding of the currently used SSL methodologies and their applications in enhancing ViTs’ capabilities. By mentioning these pretext tasks, we aim to highlight the architectures driving the SSL paradigm and its transformative impact in the field of computer vision.}
\end{figure}

\subsection{Evolution of SSL in Computer Vision}

\label{subsecBackSSL}

SSL technique is used for representation learning where a model is designed to learn the underlying relationship between its inputs. The framework of Energy-Based Models (EBMs) \citep{lecun2006tutorial} can effectively describe this objective, where the aim is to assign high energy to inputs that are incompatible and low energy to those that are compatible.
The foundation for many self-supervised learning techniques applied to images was laid by Geoffrey Hinton and others in the 1980s through the development of autoencoders and the proposal of greedy layer-wise pretraining \citep{bengio2006greedy}, where layers of a deep neural network is trained sequentially. However,  autoencoders were basically used for learning features and reducing dimensionality. They were impactful at that time as they made it possible to train the first “deep” networks. An analogous approach was Restricted Boltzmann Machines (RBMs), that allows layer-wise training and then combine those layers to build deep belief nets \citep{salakhutdinov2009deep}. Although these techniques were eventually replaced by simpler initialization strategies and extended training procedures, they played a vital part in the development of early deep learning architectures.\par

In 2000s, types of unsupervised learning, including sparse coding and K-means, were employed for self-supervised learning (SSL). These techniques aimed to learn image features without labeled data, laying the groundwork for more complex self-supervised methods. The category of Spatial Context prediction gained attention in SSL for the training of CNNs. In this regard, Pathak et al. (2016) \citep{pathak2016context} proposed inpainting using context prediction as a pretext task for SSL in images, where the model predicts missing parts of an image. Zhang et al. (2016) \citep{zhang2016colorful} explored a straightforward method of the colorization task for self-supervised representation learning. They trained a model to colorize grayscale images, which significantly enhances its capability to understand textures, patterns, and object identities. Noroozi and Favaro (2016) \citep{noroozi2016unsupervised} introduced a method where the model learns to solve jigsaw puzzles, using the relative positions of image patches as supervision. Gidaris et al. (2018) \citep{gidaris2018unsupervised} introduced RotaNet, which employs a pretext task where the model predicts the rotation angle applied to an image, enhancing feature learning for various orientations. Meanwhile, clustering approaches also progressed in SSL. Caron et al. (2018) \citep{caron2018deep} proposed DeepCluster \citep{caron2018deep}, a method that iteratively clusters image features and trains a neural network to predict the cluster assignment.

Contrastive learning became a significant approach in self-supervised learning during the era of CNNs \citep{he2020momentum, chen2020simple}. The roots of contrastive learning can be found in early work on metric learning and the development of Siamese networks \citep{bromley1993signature} in the 1990s and early 2000s. Siamese Network architecture, introduced by Bromley et al. in 1993 \citep{bromley1993signature}, involved training twin networks to lessen the relationship among embeddings of matching pairs and expanding the gap between embeddings of constrasting pairs. The formalization of contrastive loss by Hadsell, Chopra, and LeCun (2006) \citep{hadsell2006dimensionality} was a significant step forward.

In the late 2010s, the concept of employing contrastive learning for self-supervised learning gained attraction. This period saw the development of several influential techniques that leveraged contrastive objectives to learn meaningful patterns from unlabeled data. Dosovitskiy (2016) \citep{alexey2016discriminative} used instance discrimination as a pretext task. Each image was handled as a different class, and the network was trained to differentiate between differnt variations of the same image. Wu (2018). \citep{wu2018unsupervised} refined the instance discrimination approach using a memory repository to save negative examples for the contrastive loss. Oord (2018). \citep{oord2018representation} introduced method known as Contrastive Predictive Coding in 2018, which learns patterns by predicting future observations in latent space using contrastive learning. Chen (2020) \citep{chen2020simple} introduced a framework known as SimCLR for contrastive learning of visual features that trains by contrasting similar and non-similar pairs of data. This significantly advanced the performance of self-supervised methods in computer vision. Over time, the idea of hybrid learning gained strength and in 2020, Caron (2020) \citep{caron2020unsupervised} introduced SwAV, which combines contrastive learning with clustering by using swapped assignments between different augmentations of an image.\par

The idea of self-distillation gives a new dimension to SSL. Grill (2020) \citep{grill2020bootstrap} boosted the idea of SSL by introducing BYOL, a self-supervised method that avoids the need for negative samples (simplifying contrastive learning) by using two networks to predict the outputs of each other. He et al. \citep{he2020momentum} introduced MoCo, which extends the idea of SimCLR by using a dynamic dictionary and a momentum encoder for contrastive learning. Facebook AI Research (Caron \citep{caron2021emerging}) presented DINO, which uses self-distillation with no labels, leveraging VITs for SSL.\par

In 2020, the idea of transformers in vision was introduced. Vision transformers are high-capacity models and need a huge data for model tuning and good generalization. For improving the learning ability of the transformers, Dosovitskiy (2020) \citep{dosovitskiy2020image} adapted SSL techniques like masked image modeling and contrastive learning for ViT. Inspired by BERT (2018) \citep{devlin2018bert}, Dosovitskiy et al. \citep{dosovitskiy2020image} masked certain segment tokens and replaced them with trained mask tokens. Later the model was further trained for predicting pixel values directly. However, they found unsupervised pre-training technique notably less useful than pre-training with supervised methods. Before ViTs, the same idea of inpainting was employed in CNNs, where portions of an image were masked out and the model was taught to reconstruct them, this approach to pre-training is termed as MIM (Masked Image Modeling). In ViTs, the MIM was framed as a regression task. This involved initially utilizing an autoencoder, which encode image segments to separate tokens, subsequently, the transformer is pre-trained to estimate the distinct token values for masked tokens. At that time, Bert pre-training of image transformers (BEiT) \citep{bao2106beit} demonstrated substantial enhanced results in image classification and semantic segmentation compared to earlier supervised and self-supervised methods. However, its training process was intricate as it required a robust autoencoder to transform image patches into discrete tokens.\par

The SSL techniques based on cross-covariance (correlation) analysis was introduced in 2021. This family originates from the Canonical Correlation Analysis (CCA) framework developed by Hotelling in 1992 \citep{thompson2000canonical}. The main idea behind utilizing CCA is to determine the correlation among two variables by examining their cross-covariance matrices. SSL methods based on this idea include VICReg \citep{bardes2022vicreg}, Barlow Twins \citep{zbontar2021barlow}, SWAV \citep{caron2020unsupervised}, and W-MSE \citep{ermolov2021whitening}.
The latest of these methods is VICReg that balances three goals: variance, invariance, and covariance. Controlling the variance along each dimension of the representation helps to prevent collapse, ensuring that the representation maintains diversity. Invariance ensures that two views of the same data are encoded similarly, while covariance promotes various dimensions of the representation to learn distinct features. This balanced approach allows VICReg to effectively leverage cross-covariance matrices for improved self-supervised learning.\par

Recently, the idea of multi-modality gained attention with the introduction of CLIP by OpenAI in 2021 \citep{radford2021learning}. CLIP learns image and text embeddings simultaneously by matching corresponding image-caption pairs. Similarly, ALIGN \citep{jia2021scaling} uses self-supervised learning approaches to develop a shared feature space for images and their corresponding captions.
The concept of MIM pre-training was streamlined by two concurrent works: MAE by He et al. \citep{he2022masked} and SimMIM by Xie et al. \citep{xie2022simmim}. These approaches propose simplified algorithms that immediately rebuild  masked parts of image patches instead of using separate image pieces obtained from an encoder as done in BEiT \citep{bao2106beit}. This idea was inspired by masked language modeling in NLP. These straightforward pre-training strategies performs better as compared to BEiT on classifying images, object detection and semantic segmentation (downstream tasks). MIM has recently dominated this domain by obtaining state-of-the-art performance on ViTs. The essence of this technique is to enhance the network's capability to capture visual context at patch level using a denoising auto-encoding mechanism. Invariant Joint Embedding Predictive Architecture (IJEPA) \citep{assran2023self} has emerged as a promising alternative, addressing some of the limitations inherent in other self-supervised learning methods. Unlike traditional self-supervised approaches that depend heavily on hand-crafted data augmentations, IJEPA focuses on invariant feature learning and joint embeddings. IJEPA relies on predicting the patterns of destination blocks within an image from a single context block, using a strategic masking method to guide the learning process. 


\renewcommand{\arraystretch}{3}
\begin{table*}[]
\tiny
\centering
\caption{Architectural specifications of the leading SSL based models, their parameters, and performance on benchmark datasets.}
\label{tab:Pretext-Tasks}

\begin{tabular}{|p{3.3cm}|p{5cm}|p{3.7cm}|}
    \hline
    \multicolumn{1}{|l|}{\normalsize{\textbf{Pretext Task}}} & \multicolumn{1}{l|}{\normalsize{\textbf{Effective For}}} & \multicolumn{1}{l|}{\normalsize{\textbf{Ineffective For}}}   \\ \hline
    \small{RotNet \citep{gidaris2018unsupervised}}                      & \small{Spatial awareness, Object recognition}          & \small{Complex texture understanding}   \\ \hline
\small{Jigsaw Puzzles} \citep{noroozi2016unsupervised}            & \small{Object parts, Spatial layout}                   & \small{Temporal dynamics}               \\ \hline
\small{Patch Position Prediction} \citep{caron2020unsupervised}   & \small{Spatial hierarchies}                            & \small{Color/texture learning}          \\ \hline
\small{Colorization} \citep{zhang2016colorful}               & \small{Textures, Patterns, Object ID}                  & \small{Spatial relationships}           \\ \hline
\small{Inpainting}  \citep{pathak2016context}                & \small{Texture continuity, Object completion}          & \small{Temporal understanding}          \\ \hline
\small{Frame Prediction} \citep{villegas2017decomposing}           & \small{Motion patterns, Temporal consistency}          & \small{Static image features}           \\ \hline
\small{Temporal Order Verification} \citep{misra2016shuffle} & \small{Temporal understanding}                         & \small{Static spatial understanding}    \\ \hline
\small{SimCLR, MoCo, CPC} \citep{chen2020simple,he2020momentum,oord2018representation}          & \small{Feature discrimination}                         & \small{Temporal, Cross-modal learning}  \\ \hline
\small{DeepCluster, SwAV} \citep{caron2018deep,caron2020unsupervised}          & \small{Feature grouping, Discrimination}               & \small{Dynamic scene understanding}     \\ \hline
\small{BYOL, DINO, SimSiam} \citep{grill2020bootstrap,caron2021emerging,chen2021exploring}        & \small{Consistent feature learning}                    & \small{High-frequency detail understanding} \\ \hline
\small{Barlow Twins, VICReg} \citep{zbontar2021barlow,bardes2022vicreg}       & \small{Unique feature emphasis}                        & \small{Temporal feature learning}           \\ \hline
\small{Audio-Visual Correspondence} \citep{arandjelovic2017look} & \small{Cross-modal learning}                           & \small{Isolated visual feature learning}    \\ \hline
\small{Self-Labelling} \citep{asano2019self}             & \small{Simultaneous clustering and representation}     & \small{Limited scalability}             \\ \hline
\small{Local Aggregation} \citep{zhuang2019local}           & \small{Localized feature learning}                     & \small{Global contextual understanding} \\ \hline
\small{TiCo} \citep{zhu2022tico}                        & \small{Transformation invariance, Covariance contrast} & \small{Isolated feature learning}       \\ \hline
\small{Contrastive Multiview Coding} \citep{tian2020contrastive}       & \small{Multiview feature learning}    & \small{Single modality data processing}     \\ \hline
\small{Relative Predictive Coding } \citep{doersch2017multi}        & \small{Relative feature prediction}   & \small{Absolute feature understanding}      \\ \hline
\end{tabular}
\end{table*}

\subsection{Comparison of SSL with Transfer Learning}


In computer vision, transfer learning (TL) \citep{yosinski2014transferable} is employed in the development of CNN models to solve a new challenging problem by utilizing the pre-trained model's weight sharing mechanism especially in the presence of data scarcity \citep{hassan2022drug}.
Initially, transfer learning was the dominant approach, but recently, SSL has demonstrated promising results across various applications.

\begin{figure}[H]
      \centering
      \includegraphics[height=6cm, width=12cm]{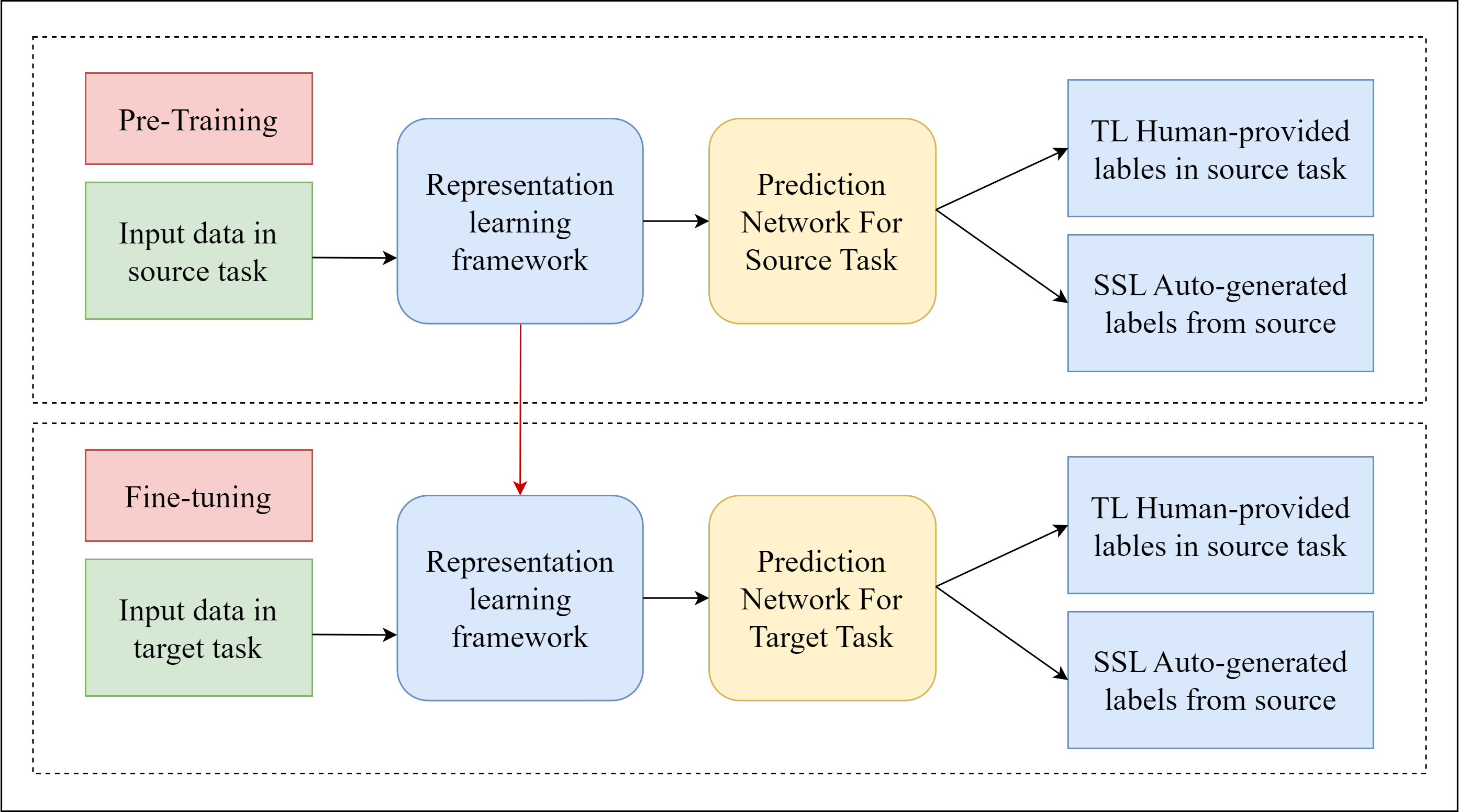}
      \caption{TL and SSL workflow.}
      \label{TL}
\end{figure}

Although TL and SSL are not mutually exclusive techniques, they have some major differences. Fig.\ref{TL} shows the TL and SSL workflow. TL and SL have two main steps: (i) training on a source task and (ii) adapting to a specific target task. The aiming of first step is to optimize the network's weight parameters to obtain a better starting point.

The challenges associated with TL or SSL especially in complex vision problems are due to several factors, including  
the volume of pre-training data and class imbalances in the source task. A study by \cite{yang2020transfer} examined this issue across multiple sources and target tasks, yielding the following key insights:
 \begin{itemize}
     \item {When the visual scenarios in the training and target applications are significantly different—say, moving from natural scenes to medical images— SSL tends to perform better. On the other hand, TL is more effective when both the training and target tasks involve similar visual elements.}

    \item{Working with a limited set of images for pre-training SSL offers better performance. However, availability of massive dataset for pre-training TL seems to outperform SSL.}

    \item{SSL offers a more robust performance in transitioning between various types of visual tasks. For example, a model trained using SSL might adapt more easily from facial recognition to gesture recognition. Meanwhile TL may offers better performance when switching between these tasks \citep{ericsson2021well}.}

    \item{When both the training and target tasks involve similar types of images, TL performance can vary widely. Whereas, SSL remains more stable under these conditions.}

    \item{In class imbalance problem especially in vision tasks, SSL shows resilient compared to TL.}

    \item{In SSL, supplementing pre-training with some of the target task's data, for instance, using some annotated images/data can enhance the model performance. This strategy is not as effective for TL.}
 \end{itemize}
 It is concluded that there is a trade-offs between these two methods and should be carefully considered according to specific needs and constraints of the vision task at hand. Emerging research continues to provide valuable insights into these methodologies' comparative strengths and weaknesses, guiding practitioners in making informed choices for their specific use cases.
\begin{table*}[]
\centering
\caption{Comparative summary of transfer learning and self-supervised learning (SSL) based on various aspects including task dependence, data requirements, and computational overhead}
\label{tab:Comparative summary of transfer learning and self-supervised learning}
\renewcommand{\arraystretch}{2.0}  
\begin{tabular}{|p{2cm}|p{4.8cm}|p{5.5cm}|}
\hline
\textbf{Aspects} & \textbf{Transfer Learning} & \textbf{SSL} \\ \hline
Pre-training and Fine-tuning &
  \begin{tabular}[c]{@{}l@{}}Pre-training on a source task \\first and then fine-tuning on a \\ target task \citep{yosinski2014transferable} \end{tabular} &
  \begin{tabular}[c]{@{}l@{}}Directly trained on an auxiliary task \\ created from the available data \\ \citep{doersch2015unsupervised} \end{tabular} \\ \hline
Task Dependence &
  \begin{tabular}[c]{@{}l@{}}Presumes that the source and \\ target tasks are correlated \\ \citep{yang2020transfer} \end{tabular} &
  \begin{tabular}[c]{@{}l@{}}Does not assume any particular task; \\ aims for generalized feature learning \\ \citep{singh2018self} \end{tabular} \\ \hline
Data Requirements &
  \begin{tabular}[c]{@{}l@{}}Works well with few labeled \\ examples if source task data is \\ similar \citep{raina2007self} \end{tabular} &
  \begin{tabular}[c]{@{}l@{}}Needs a huge amount of unlabeled \\ data for feature learning. Afterwards \\ fine-tuned on a relatively smaller \\ labeled dataset \citep{chen2020simple} \end{tabular} \\ \hline
Flexibility &
  Limited to the domain of the source task \citep{yang2020transfer} &
  \begin{tabular}[c]{@{}l@{}}More flexible; aims to learn features \\ useful for a wide range of tasks \\ \citep{singh2018self}\end{tabular} \\ \hline
Data Efficiency &  \begin{tabular}[c]{@{}l@{}}More data-efficient when a \\closely related source task is \\available \citep{raina2007self} \end{tabular} &
  \begin{tabular}[c]{@{}l@{}}May require more data for the initial \\ learning phase, but becomes efficient \\ when the learned features are applied to \\ multiple tasks \citep{chen2020simple}\end{tabular} \\ \hline
Computational Overhead &
  \begin{tabular}[c]{@{}l@{}}Generally lower, especially \\during the fine-tuning phase \\ \citep{yosinski2014transferable} \end{tabular} &
  \begin{tabular}[c]{@{}l@{}}Usually higher during the initial training \\ phase to learn from unlabeled data \\\citep{he2020momentum} \end{tabular} \\ \hline
End Goal & 
  \begin{tabular}[c]{@{}l@{}}The end goal (e.g., specific \\ classification  task) is often known \\ beforehand \citep{oquab2014learning} \end{tabular} &
  \begin{tabular}[c]{@{}l@{}}Intension is to learn generalized features  \\ that can be applied to various downstream \\ tasks \citep{he2020momentum} \end{tabular} \\ \hline
\end{tabular}
\end{table*}

\section{SSL based Techniques for ViTs}\label{sec3}
In computer vision, the success of ViTs is based on large-scale self-supervised pre-training. This paper categorizes the approaches for self-supervised pre-training in ViTs into five groups: Contrastive, Generative, Clustering, knowledge distillation and Hybrid SSL methods. Each methodology's nuances and impact on transformer learning are discussed in detail in subsequent sections.


\subsection{Contrastive Methods}
Contrastive Learning (CL) stands as a prominent technique in SSL \citep{el2025self, he2020momentum, chen2020simple}. Its aim is to capture invariant semantics through pairs of random views, termed Contrastive multi-view coding. CL ensures that global projections of representations are similar for positive samples and dissimilar for negative samples. This section delves into literature reviews on Contrastive Learning applied in transformer frameworks \citep{jia2025cltp}.


\textbf{Literature based Review of Contrastive Learning}

\cite{li2025domain} suggested a methodology to enhance domain generalization in mammographic image analysis using contrastive learning. Their approach consists of two stages: first, a multi-style and multi-view contrastive learning pretrainig phase, where CycleGAN and image blending techniques generate diverse mammogram styles to enhance feature robustness; second, a fine-tuning stage for downstream tasks such as mass detection, mass matching, BI-RADS classification, and breast density classification. By learning style-variant representations, their method demonstrated improved performance across multiple mammographic datasets, outperforming existing domain generalization techniques.
\cite{wang2022transformer} introduced the method SRCL that is semantically-relevant contrastive learning for histopathology image analysis. Unlike traditional contrastive learning, SRCL evaluates similarity among instances to identify additional positive pairs by arranging various positive examples that share related visual concepts. This approach enriches the diversity of positive pairs and yields more informative patterns. They employee a hybrid model named CTransPath, in which they combines a CNN with a multi-scale Swin Transformer architecture, their methodology acts as a unified feature extractor, addressing both local and global features during pre-training, seeking to acquire global feature interpretations tailored for activities within the domain of histopathology images. The efficacy of SRCL-pretrained CTransPath was evaluated across five distinct downstream tasks spanning nine publicly available datasets, such as retrieving patches, classifying patches, performing weakly-supervised classification of whole-slide images, detecting mitosis, and segmenting glands in colorectal adenocarcinoma. The findings demonstrated that the visual representations derived from SRCL not just achieved state-of-the-art performance throughout all datasets but it also exhibited increased robustness and transferability compared to both supervised and self-supervised ImageNet pre-training techniques.

To address instability issues during training, \cite{chen2021mocov3} utilized SSL framework in MOCO, representing an advancement in Contrastive Learning (CL) methodologies. MOCO utilizes dictionaries to gauge similarity between encoded keys and queries. Empirical assessments suggested that self-supervised Transformers within the MOCO framework can demonstrate good performance with relatively fewer inductive biases compared to ImageNet supervised ViT. This approach aims for a better balance between simplicity, accuracy, and scalability, showcasing its superiority over supervised pre-training methods in multiple detection and segmentation tasks. The study proposes replacing the patch projection layer in ViT with fixed random patch projections and suggests removing positional embeddings in SSL setups to address instability.

\cite{mo2023multi} in 2023 introduced MCVT, which focuses on projecting class tokens to embedding space during the initial or final phases of the ViTs using multi-layer perceptrons. In this methodology, InfoNCE loss is applied to low-level features while ProtoNCE loss is introduced in high-level features, enhancing performance in downstream classification tasks. Extensive experimentation transferring MCVT pre-trained backbones to various downstream tasks confirmed its efficacy across multiple vision-related tasks.

\cite{radford2021learning} proposed a Contrastive Language-Image pre-training (CLIP) framework, which incorporates transformers, learns image representations through natural language supervision. In pre-training, CLIP concurrently trains image and text encoders to accurately predict corresponding pairs within a batch of (image, text) training instances. CLIP discerns among the
\textit{N * N} 
possible (image, text) pairings within a batch. It achieves this by learning a multi-modal embedding space, optimizing the image and text encoders jointly to increase the cosine similarity of embeddings for the N genuine pairs while decreasing the resemblance of embeddings for the 
( N \^ 2 - N )
incorrect pairings. Symmetric cross-entropy loss is used in this optimization strategy to adjust these similarity scores. After pre-training, the model uses natural language to reference acquired visual concepts, facilitating zero-shot transfer to downstream tasks, demonstrating substantial transferability to diverse computer vision tasks without the need for task-specific training data. Evaluation encompasses benchmarking across more than 30 diverse computer vision datasets. Remarkably, the model demonstrates substantial transferability to most tasks and frequently matches or surpasses fully supervised baselines without requiring task-specific training data.

\subsubsection{Knowledge Distillation Methods}
Knowledge distillation transfers knowledge from a complex teacher model to a simpler student model without labeled data. This section explores literature on knowledge distillation in SSL applied to ViTs.

\textbf{Literature based Review of Knowledge Distillation}
\cite{caron2021emerging} in 2021 introduced knowledge distillation into SSL to enhance the training of ViTs. Their method, DINO, comprises two identical networks termed student and teacher networks, each with distinct parameters. Both networks include an encoder or base network, utilizing either ViTs or ResNet50, along with a projection head placed on top of the encoder network. The DINO approach involves generating multiple crops from two perspectives: a local perspective and a global perspective. The global crop perspective is fed to the teacher network, while the local crop perspective is fed to the student network. The local crop is a subset of the global perspective, exhibiting overlap between the two. The teacher network guides updates in the student network, but gradient stop is applied to halt backpropagation in the teacher network. Meanwhile, the teacher network weights are adjusted using the EMA (exponential moving average) of the weights from student network.

cite{li2022exploring} proposed Efficient Self-Supervised ViTs (EsViT) to achieve a optimized balance between accuracy and efficiency, reducing the cost in building state-of-the-art SSL vision systems while demonstrating enhanced scaling performance on accuracy versus throughput and model size. They introduced a multi-stage Transformer architecture, combining networks to address the issue of semantic information loss when transitioning from a single-stage transformer to a multi-stage architecture. Employing a distillation learning strategy targeting correspondence between patches, they introduced a new non-contrastive region-matching pre-training task, enhancing the model's ability to learn intricate regional dependencies and substantially improve the quality of visual representations learned. The Multi-stage ViT segments an input RGB image into individual non-overlapping patches, handling each patch as an individual token in the first stage. Later, the patch merging module concatenates the features of each group and applies a linear layer to reduce the number of tokens. Subsequently, a Transformer with a sparse self-attention module enables interactions among the merged features. This process is repeated multiple times, resulting in a hierarchical representation. EsViT, the proposed model, is validated across various tasks. EsViT achieves 81.3\% top-1 accuracy, with the linear evaluation protocol, demonstrating 3.5× parameter efficiency and at least 10× higher throughput compared to previous state-of-the-art models. EsViT leads over its supervised counterpart, Swin Transformers, by outperforming on 17 of 18 datasets in downstream linear classification tasks.

\cite{xie2021self} introduced MoBY, a SSL framework tailored for ViTs. This approach combines MoCo v2 and BYOL approaches, yielding notably high accuracy on the ImageNet-1K dataset. MoBY adopts the key queue, momentum design, and contrastive loss from MoCo v2, and it also take asymmetric data augmentations, asymmetric encoders, and momentum scheduler from BYOL. Employing a knowledge distillation approach, MoBY employees a pair of neural networks (online and target) to enhance mutual learning. Both encoders includes a backbone and a projector head, with the online encoder featuring an additional prediction head, rendering the two encoders non-uniform. Gradients progressively optimizes the online encoder, while the target encoder is a moving average of the online encoder updated through momentum. MoBY is evaluated across diverse downstream tasks, such as object detection and semantic segmentation, showcasing robustness and transferability across varied visual tasks. When utilizing DeiT-S and Swin-Transformers as backbone architectures, MoBY achieves top-1 accuracy rates of 72.8\% and 75.0\%, respectively, on ImageNet-1K.

\subsubsection{\textbf{Clustering Methods} }
Clustering-based SSL utilizes unsupervised clustering algorithms to learn representations from unlabeled data. This technique groups similar data points based on specific criteria, such as feature similarity, and then predicts these clusters or cluster assignments. By encouraging the model to distinguish between different clusters, clustering-based SSL extracts semantically rich features for downstream tasks like classification or segmentation. This approach proves valuable when labeled data is limited or costly to obtain, enabling the model to learn useful representations directly from the unlabeled data.

Clustering-based methods segment the latent space by applying clustering. This clustering can be performed offline, utilizing the entire dataset, as in  PCL \citep{li2020prototypical}, DeepCluster v2 \citep{caron2020unsupervised}, and SeLa \citep{asano2019self}, or online, using mini-batches, as in DINO( \citep{caron2021emerging} and SwAV (\citep{caron2020unsupervised}. These approaches enforce consistent clustering assignments of positive pairs using cross-entropy loss. However, previous literature have not taken advantage of this beneficial property in semi-supervised scenarios. Our study aims to fill this gap.

\textbf{Literature based Review of Clustering Methods}

\cite{hemalatha2025self} introduced a novel approach to enhance cervical cancer detection by using self-supervised learning techniques on diverse cell images.

\cite{su2024flsl} introduced a clustering-based SSL framework tailored for ViTs, aiming to improve the performance of tasks related to object detection and segmentation. Their proposed methodology, Feature-Level SSL (FLSL), works at two levels of semantics: intra-view, which deals with individual image, and inter-view, which considers over an entire dataset. FLSL utilizes a bilevel feature clustering approach, integrating mean-shift clustering intrinsic to transformers for extracting modes as patterns with a k-means-based SSL technique, ensuring semantic coherence of extracted representations both locally and globally. At the first semantic level, FLSL optimizes intra-cluster affinity using a self-attention layer to encourage semantic representations within clusters. On the second tier, semantic representations are cultivated through non-empty k-means clustering, with positive samples identified using a cross-attention layer. Experimental outcomes illustrate the efficacy of FLSL in dense prediction tasks, achieving significant improvements in object detection and instance-level segmentation on MS-COCO, utilizing Mask R-CNN with ViT-S/16 and ViT-S/8 as backbones, respectively.

\cite{fini2023semi} leverage the limited annotations in the semi-supervised setting to enhance the quality of learned representations. The core idea involves substituting cluster centroids with class prototypes learned through supervised methods, guiding unlabeled data to cluster around these prototypes through a self-supervised clustering-based objective.
Their approach involves the joint optimization of a supervised loss on labeled data and a self-supervised loss on unlabeled data.
For a given an unannotated dataset 
\textit{Ki = \{x\textsubscript{1}i , ..., x\textsubscript{N}i \}}, 
two versions of each input image, (x, xˆ), are created through data augmentation technique. These versions are then processed through an encoder network f\( \theta \). This encoder network is comprised of a projector \textit{p}, backbone \textit{b} and a set of centroids or prototypes \textit{c} implemented as a bias-free linear layer.
The forward pass through this network yields two latent demonstrations (p, pˆ), corresponding to the two augmented views. The prototypes \textit{c} are then employed to produce logits (c, cˆ) for each representations. These logits are transformed into probability distributions over clusters using softmax function $\sigma$ resulting in cluster assignments. 

\subsubsection{\textbf{Generative Methods}}
Generative approaches, for example Generative Adversarial Networks (GANs) or Autoencoders, provide a promising avenue for pre-training ViTs. These models not only generate realistic images or reconstruct input images, but also encode meaningful representations of visual content, facilitating enhanced learning of features in subsequent tasks in \citep{showrov2024generative}.

\textbf{Literature based Review of Generative Methods}

\cite{el2025self} suggest that both contrastive learning and in-painting are effective auxiliary tasks for SSL in the medical domain, leading to a 5.26\% increase in accuracy for polyp classification compared to other weight initialization methods.

\cite{lin2025self} proposed a self-supervised BM-GAN to learn morphological changes in MRI over the course of a disease. BM-GAN, learns both forward and backward temporal dependencies to interpolate missing time-series data at arbitrary time points. By integrating self-supervised learning and a bidirectional GAN framework, the model ensures realistic reconstruction without labeled data, offering a robust solution for handling irregular and incomplete time-series datasets.

\cite{chen2020big} introduced iGPT, an image-based version of GPT, for self-supervised visual learning. Unlike ViT, which uses patch embedding, iGPT directly resizes and flattens images into lower-resolution sequences, which are then fed into a GPT-2 \citep{radford2019language} model for autoregressive pixel prediction. Despite its significant computational cost, iGPT achieves high accuracy in CIFAR-10, surpassing supervised wide ResNet.

\cite{bao2021beit} introduced BEIT, that is, bidirectional encoder representation from image transformers is a self-supervised vision representation model. Instead of pixel-wise generation, they adopted MIM as a task to pretrain ViTs in a self-supervised manner. BEIT, grounded on a BERT-style visual transformer, reconstructs masked images in the latent space. The approach involves randomly masking parts of image pieces and predicting the visual tokens relevant to these hidden pieces. The main motivation behind this pre-training technique is to restore the base visual tokens utilizing corrupted image patches. The standard visual Transformer serves as the backbone network. Evaluation using thorough fine-tuning for tasks such as image classification and semantic segmentation demonstrated competitive performance without human annotation.

\cite{he2022masked} proposed MAE (Masked Autoencoders), in this methodology parts of input images are randomly masked and subsequently reconstructed the pixels that were masked, enabling optimized training of large models that demonstrate strong generalize across downstream tasks. This method utilizes an asymmetric encoder-decoder setup, where the encoder only looks at the visible parts of the image, and a small decoder rebuilds the original image using the learned representation and mask tokens. By hiding a large proportion of the input image, e.g., 75\%, MAE creates a challenging and useful self-supervised learning task. This scalable method allows large models to train efficiently and effectively, resulting in powerful models that generalize well. For instance, a standard ViT-Huge model achieved the highest accuracy of (87.8\%) among techniques using only ImageNet-1K data. Supervised pre-training was outperformed by transfer learning in downstream task, showing strong potential for scaling behavior.

\subsubsection{Hybrid Methods}
Hybrid methods similar to generative approaches often combine the strengths of different techniques to achieve specific objectives \citep{shao2024hybrid}.

\textbf{Literature based Review of Hybrid Methods}

\citep{zhao2025maemc}, proposed a hybrid deep learning model MAEMC-NET, which combines masked autoEncoders for reconstructing masked CT patches and Momentum Contrast (MoCo) for learning discriminative features from different CT views.

\citep{wu2024eslaxdet} implemented the hybrid SSL pre-training framework Enhanced SSL with Masked Autoencoders (ESLA). This innovative approach integrates both contrastive learning (CL) and MAE frameworks to enhance SSL. ESLA improves model generalization and representation while addressing issues stemming from competing features during data enhancement. This is achieved through a masking mechanism in the CL framework and by blending representations of different views in the latent space. 

During training, ESLA conducts data augmentation and masking on input data to create two views. 
The online encoder receives EM$\theta$ and target encoder get EM.
The "mixer" combines \textit{latent1} with \textit{latent2} and restoration masks to generate a "mixed latent," which is then used to compute the reconstruction loss. ESLA's structure comprise of two Interconnected branches known as CL and MAE branch respectively. There are two kinds of encoders in CL branch including  online and target encoder that generate distinct intermediate representations (IRs), \textit{latent1} and \textit{latent2}. To decrease the computational power, an encoder incorporating a masking technique is used to encode two modified variants, referring to the same image. 
A contrast loss is generated by evaluating the \textit{latent1} and \textit{latent2} using predictor and projector.
Rather than being trained with gradient descent, the target network uses an exact replica of online  network,
following it with a delayed using an EMA that is exponential moving average. 
In MAE branch, a feature map is generated by combining the information with the help of mixer that corresponds to orignal image.
This feature map is unmasked and forwarded to the decoder for computing the reconstruction loss. In this context shallow data enhancement occur in pixel space (\textit{D}) that produces two distinct views of the same image while in the latent space {\textit{L}} deep data enhancement is executed that generate the "mixed latent", as an aspect of the decoder input. 

\cite{quan2022global} introduced a novel SSL model, based on the global contrast-masked autoencoder (GCMAE), which adeptly extracts features from pathological images including both local and global features. GCMAE leverages masking image reconstruction and contrastive learning as self-supervised pretext tasks, enabling the encoder part to proficiently depict local and global features. This model is comprises of an initializer, encoder, a tile feature extractor, and a global feature extractor facilitating image reconstruction and contrastive learning tasks. In this investigation, an asymmetric encoder-decoder architecture is employed. ViT serves as the backbone for the encoder, pretrained using MAE. The module responsible for extracting features from patches consists of 8 transformer blocks, creating an asymmetric configuration when combined with an encoder comprising approximately 12 transformer blocks, as seen in the ViT-base architecture.
This asymmetry facilitates decoupling of the encoder and decoder, fostering the acquisition of more generalized representations. Extraction of global feature is achieved via utilizing contrastive learning. The hidden representation \textit{V}patch(\textit{vis}) is used to reconstruct image through decoder and update a memory bank (\textit{B}). This updation of \textit{B} is executed to store the global features utilizing momentum coefficient \textit{t}. The dynamically update is conducted to pursue the feature vectors of image data using memory bank as a fixed size queue. The features randomly selected from memory bank are utilized as negative samples, that helps to reduce the effects of the constraint caused by batch size on the heavy performance
Notably, the momentum adjustment mechanism is designed distinctively from MoCo's model parameter update, it employ an encoder structure and a memory bank to facilitate contrastive learning. 
Vpath(vis) is depicted as the encoders output in the current epoch in the latent representation, where as the vpatch(vis) represented the previous epoch latent representation stored in \textit{B}.
Particularly, the latent demonstrations at the output of encoder as \textit{v}patch(vis) in the current epoch,
The cost function combines both, the weighted sum of MSE loss for tile feature extraction and the NCE loss for global feature extraction. This formulation aims to minimize the gap between similar features, enhancing the model's generalization and accuracy, particularly in cross-dataset transfer learning tasks.

\cite{fang2023hybrid} proposed a hybrid SSL framework (HSL) for Hyperspectral Image (HSI) classification. Leveraging MIM and contrastive learning to enhance its effectiveness. Employing a ViT as the backbone network within an asymmetric encoder-decoder architecture with two branches, 
HSL achieve improved HSI classification utilizing contrastive pre-training and self-supervised generative tasks by effectively extracting spatial and spectral details.
To facilitate image reconstruction through masked inputs, each branch connects intermediate features of the encoder to the decoder at the corresponding level.
Contrastive learning is accomplished by utilizing the intermediate features that are obtained from two encoders that share parameters across both branches. Utilizing standard data augmentation methods, HSL generates several views of input images. Then each image view undergoes random masking technique such as MAE. 
Initially, image segments with mask are reorganised into a series of patches and later processed by HSL encoder and decoder to extract patterns. Extracted features are then pass through an adaptive average pooling layer and a projection head to achieve the contrastive learning task of capturing image resemblance. The decoder, which also includes multiple Transformer encoders, rebuilds patches after certain patches have been masked. Moreover, the Transformer Encoder retains the initial structure while the two output features from the encoder are used to compute contrastive loss. The loss is calculated using the decoded sequence produced by the decoder and the sequence that precedes the encoder masking, efficiently completing the masked image rebulding process.
In summary, HSL combines contrastive pre-training and self-supervised generative tasks, facilitating effectiveness for downstream HSI classification task.

\section{Regularization Techniques}

In basic terminology, regularization techniques helps in making model simpler. As Occam's razor principle state, simpler models usually work better, this means keeping things straightforward is key. By using different techniques, we narrow down the model's options to make it more focused. The integration of SSL methods with traditional regularization techniques offers a nuanced approach to improving the generalizability and robustness of vision transformer models. This analysis explores how various SSL methods incorporate regularization principles within their frameworks.

\subsection{BYOL}
Bootstrap Your Own Latent (BYOL) \citep{grill2020bootstrap} employs a teacher-student framework, where the student network predicts the output of a teacher network on augmented inputs. The teacher network is updated as a moving average of the student's parameters, ensuring learning consistency. This method mirrors the stabilization effect seen in regularization techniques like batch normalization and parameter sharing.

\subsection{MoCo}
Momentum Contrast (MoCo) \citep{he2020momentum} leverages a momentum-based moving average of the query encoder, establishing a large and consistent dictionary. This approach acts as a dynamic regularization mechanism, ensuring stability in the representation of dictionary keys over time. This stability is vital for the extraction of stable and generalizable features, akin to how traditional regularization methods prevent overfitting by maintaining consistency in model parameters.

\subsection{SimCLR}
SimCLR \citep{chen2020simple}  utilizes a straightforward contrastive learning framework, where its regularization is primarily through aggressive data augmentation techniques. These augmentations consist of random cropping, resizing, color manipulation, and the application of Gaussian blur, enabling the model to learn invariant and robust features. This is similar to how techniques like dropout and data augmentation in traditional settings prevent over-reliance on specific input features.

\subsection{DINO}
SINO (Self-Distillation with No Labels) \citep{caron2021emerging} utilizes self-distillation for regularization, where the student network mimics the teacher network, updated as an exponential moving average of the student's parameters. This method is reminiscent of techniques like early stopping and weight decay, which aim to prevent overfitting by smoothing the learning trajectory.

\subsection{Barlow Twins}
Barlow Twins \citep{zbontar2021barlow}
method aims to align the cross-correlation matrix between the outputs of twin networks with the identity matrix. This alignment encourages the model to learn features invariant to input distortions, functioning as a regularization technique by discouraging the learning of redundant features.

\subsection{SwAV}
SwAV (Swapping Assignments between Views) \citep{caron2020unsupervised}  regularization is unique, involving clustering approaches and consistency enforcement between cluster assignments for alternate projections of the single image. This approach is akin to enforcing a constraint (regularization) on the feature space to enhance generalizability.

\subsection{TiCo}
TiCo (Time-Contrastive Consistent Learning of Visual Representations) \citep{zhu2022tico}
focuses on learning temporally consistent representations, acting as a regularization mechanism by ensuring robustness to temporal variations in data. This is similar to entropy regularization, which encourages the model to learn more uncertain, hence generalizable, features.

\subsection{CMC}
CMC (Contrastive Multiview Coding) \citep{tian2020contrastive} regularizes learning by enforcing invariance to different views of the same scene. This invariance is a form of regularization, similar to dropout variations that encourage the model to learn features that are invariant to specific changes in the input.

\subsection{SELF-LABELLING}
SELF-LABELLING \citep{asano2019self} approach involves pseudo-labeling, where the network assigns labels to its inputs and learns from these self-generated labels. This process regularizes the model by forcing it to generalize from its initial, potentially noisy, self-generated labels, akin to label smoothing which also deals with label noise.

In conclusion, these SSL methods embody various facets of regularization, each employing unique strategies to instill robustness and generalization in the learned representations. They extend the traditional notion of regularization beyond mere parameter adjustment, delving into feature space manipulation, consistency enforcement, and self-reflection in learning processes.
\section{Evaluation Metrics, Benchmarks and loss functions}\label{sec4}

In this section, we will discuss the evaluation metrics and benchmarks used to evaluate the performance of SSL (SSL) methods in ViTs.

\subsection{Performance Metrics for SSL in ViTs}
%
We analyze the general representation quality and task-specific effectiveness of various performance metrics to evaluate self-supervised learning (SSL) in Vision Transformers (ViTs). In addition to few and zero shots where little or no label is available during test time, we consider classification Top-1 accuracy, Average Precision (AP) for detection, and mean Intersection over Union (mIoU) for segmentation as the most important metrics.

\subsubsection{Accuracy}
\begin{equation}
    Accuracy =  \frac{True Positives + True Negatives}{True Positives + True Negatives + False Positives + False Negatives}
\end{equation}

It may be noted that accuracy is generally a good metric as long as the datasets are well-balanced. However, with imbalanced classes, it may shows erroneous readings as the numbers of rightly classified samples of different classes are not considered.

Precision, also known as a measure of quality, quantifies the portion of correctly predicted positive instances. Specifically, it is calculated as the ratio of correctly predicted positive samples to the total number of samples predicted as positive:

\begin{equation}
Precision = \frac{True Positives}{True Positives + False Positives}
\end{equation}

\subsection{Benchmark Datasets for SSL in ViTs}

Several benchmark datasets have been used to evaluate the performance of SSL methods in ViTs. Some of them are discussed below.

\subsubsection{ImageNet}

The ImageNet classification dataset \citep{ILSVRC15} contains roughly 1.2 million images for training, fifty thousand images for validation, and 100000 images for testing. Overall, the dataset is divided into one thousand diverse object classes including tablelamp, goldfish, radiator, screwdriver, and so on.

\subsubsection{COCO}
COCO (Common Objects in Context) \citep{lin2014coco} is a large-scale object detection, key-point detection, segmentation, and captioning dataset. The dataset contains over 330,000 images of which around 200,000 are fully labeled. The dataset is divided into 80 object classes and over 90 generic `stuff' classes.

\subsubsection{CIFAR-10}

The CIFAR-10 a short for Canadian Institute For Advanced Research is a dataset \citep{krizhevsky2009cifar} that contains sixty thousand 
images divided into 10 classes in which every class contains 6000 images. In total, 50000 images are utilized for training purpose while the rest of 10000 images are utilized for testing purpose. The dataset comprises of 10 classes, including horse, airplane, frog, automobile, cat, deer, bird, dog, ship, and truck.

\subsubsection{CIFAR-100}
CIFAR-100 dataset \citep{krizhevsky2009cifar} is an extension of CIFAR-10. Although the total number of images are same on both datasetss, i.e., sixty thousand, CIFAR-100 is divided into 100 distinct classes with 600 images per class. Similar to CIFAR-10, a total of 50000 images are utilized for training and 10000 images are employed for testing. Every image contains two labels, a coarse label (general category) and a fine label (particular class). Some particular object classes include apple, bee, cattle, chair, rabbit, and wowan, whereas general categories include flowers, household\_furniture, medium\_mammals, people, etc.

\subsection{Loss functions}

Machine learning algorithms must include loss functions to gauge how well a model
is doing on the job at hand and acts as a feedback signal to adjust the model’s
parameters during training. By identifying and quantifying the discrepancy between expected output and the actual target labels, they play a critical part in directing the learning process. Over the past few years, contrastive learning has appeared as an effective approach in the realm of unsupervised representation learning, revolutionizing how neural networks can be trained without labeled data. This approach hinges on the idea of learning meaningful representations by increasing the similarity among positive pairs and decreasing the similarity of negative pairs.

\subsubsection{InfoNCE Loss}
InfoNCE loss (Information Noise Contrastive Estimation) \citep{oord2018representation} is a loss function commonly used in contrastive learning, particularly in self-supervised learning tasks. It operates by introducing noise into the input data to create negative pairs, enabling the model to differentiate between real positive pairs and artificially generated negatives.

Given a query $q$ and a set of keys ${k_0, k_1, \dots, k_N}$, where $k_0$ is the positive key and $k_1, \dots, k_N$ are negative keys, the InfoNCE loss is formulated as:

\begin{equation}
\mathcal{L} = - \mathbb{E} \left[ \log \frac{\exp(s(q, k_0) / \tau)}{\sum_{j=0}^{N} \exp(s(q, k_j) / \tau)} \right]
\end{equation}

\subsubsection{NT-Xent/SimCLR Loss}

NT-Xent (Normalized Temperature Scaled Cross Entropy)  \citep{chen2020simple}, which extends InfoNCE by incorporating a temperature parameter to scale the pairwise similarities. Through a softmax function, it computes the negative log-likelihood of positive pairs, enhancing the overall contrastive learning process. 

The NT-Xent loss is defined for a pair of augmented views $(i, j)$ of the same instance in a batch. First, the similarity between embeddings $z_i$ and $z_j$ is computed using cosine similarity:

\begin{equation}
sim(z_i, z_j) = \frac{z_i \cdot z_j}{|z_i| |z_j|}
\end{equation}

The loss function for a positive pair $(i, j)$ is given as:

\begin{equation}
\mathcal{L}{i,j} = - \log \frac{\exp(sim(z_i, z_j)/\tau)}{\sum\limits{k \neq i} \exp(sim(z_i, z_k)/\tau)}
\end{equation}

\subsubsection{Triplet Loss} 

In many machine learning tasks, particularly in face recognition, person identification, and similarity-based retrieval, a key challenge is learning meaningful feature representations that ensure similar instances are clustered together while dissimilar ones are well separated. Traditional classification-based losses struggle in scenarios with high intraclass variability and low interclass differences, leading to misclassifications. Triplet loss \citep{schroff2015facenet} addresses this issue by using a relative distance-based approach, which ensures that an anchor sample is closer to a positive sample (the same class) than to a negative sample (different class) by a specified margin. By simultaneously minimizing the anchor-positive distance and maximizing the anchor-negative distance, Triplet Loss enforces a structured embedding space that enhances discrimination. This approach has proven highly effective in deep metric learning tasks, such as FaceNet, where learning compact and well-separated embeddings is crucial for robust face verification, clustering, and retrieval in unseen data scenarios.

Given an anchor sample $x_a$, a positive sample $x_p$, and a negative sample $x_n$, the triplet loss is formulated as follows:

\begin{equation}
\mathcal{L} = \sum_{i} \max \left( d(f(x_a^i), f(x_p^i)) - d(f(x_a^i), f(x_n^i)) + \alpha, 0 \right)
\end{equation}

\subsubsection{Barlow Twins Loss}

In self-supervised learning, achieving meaningful feature representations without labeled data is challenging, especially when avoiding collapse, where embeddings become trivial or redundant. Unlike contrastive approaches that rely on distinguishing between different samples, Barlow Twins Loss takes a redundancy reduction approach to ensure diverse and informative features. It aligns representations of augmented views of the same sample, while reducing unnecessary correlations across feature dimensions. This is done by computing the cross-correlation matrix between embeddings from two views and encouraging it to be as close to the identity matrix. By minimizing redundancy without relying on negative pairs, Barlow Twins enables models to learn robust and generalizable representations, making it effective for tasks where contrastive learning may be limited by batch size constraints or reliance on negative samples.

Given two embeddings $Z_A$ and $Z_B$ corresponding to two different augmentations of the same sample, we define the cross-correlation matrix $C$ as:

\begin{equation}
C_{ij} = \frac{\sum_{b} z_{A, b}^{i} z_{B, b}^{j}}{N}
\end{equation}

where $N$ is the batch size, and $z_{A, b}^{i}$ represents the $i$-th feature of the embedding $Z_A$ for the sample $b$.

The Barlow Twins Loss is then formulated as:

\begin{equation}
\mathcal{L} = \sum_{i} (1 - C_{ii})^2 + \lambda \sum_{i \neq j} C_{ij}^2
\end{equation}

\subsubsection{Reconstruction Loss - MSE - L1 Loss}


Reconstruction loss is commonly used in tasks such as autoencoders and generative models to measure how well a reconstructed output matches the original input.

\textbf{Mean Squared Error (MSE) Loss}

MSE computes the squared difference between predicted and actual values:

\begin{equation}
\mathcal{L}_{MSE} = \frac{1}{N} \sum_{i=1}^{N} (y_i - \hat{y}_i)^2
\end{equation}

MSE penalizes large errors more due to the squared term, making it sensitive to outliers.

\textbf{L1 Loss (Mean Absolute Error, MAE)}

L1 loss calculates the absolute difference between predicted and actual values:

\begin{equation}
\mathcal{L}_{L1} = \frac{1}{N} \sum_{i=1}^{N} |y_i - \hat{y}_i|
\end{equation}

Unlike MSE, L1 Loss is more robust to outliers as it does not square the errors. It is often used when preserving small details in reconstructions is important.

\subsubsection{Circle Loss}

In response to the shortcomings of triplet loss, the Circle Loss\citep{sun2020circle} was recently developed. By establishing a decision boundary in a hypersphere around instances and incorporating a flexible buffer based on the instance distances, it encourages positive pairs to lie within the boundary while keeping negative pairs outside.

These are just a few instances of the wide array of contrastive loss functions employed in various research works. The selection of a loss function is determined by factors like the specific problem, the nature of the data, and the required characteristics of the learned representations. Researchers often experiment with multiple loss functions to determine the most suitable one for their specific task \citep{wang2020comprehensive}. As contrastive learning progresses, more innovative loss formulations and improvements are likely to emerge, opening up new possibilities for unsupervised representation learning .

\subsection{Comparison with Supervised Learning Approaches}

SSL, particularly in the context of ViTs, has completely changed the area of computer vision. Large volumes of unlabeled data can be used with these techniques to acquire meaningful representations, reducing the substantial reliance on labeled instances observed in conventional supervised learning. The main benefit of SSL is its capacity to take advantage of the wealth of unlabeled data that is readily available online \citep{chen2020simple}, allowing models to learn from big datasets without the need for explicit annotations. SSL is especially appealing for real-world applications because of its scalability and affordability given that manual data labeling can be resource-intensive, costly, and require significant effort \citep{gui2024survey}.

The comparative evaluation of SSL and supervised learning methods using the same downstream tasks and evaluation metrics has shed light on the effectiveness of SSL in ViTs. In situations where the annotated data is limited or challenging to gather, SSL has demonstrated its potential to achieve competitive or even superior performance. This is especially valuable for real-world scenarios where manually annotating data might not be feasible, yet high-performance computer vision models are needed. A comprehensive learning framework known as "SSL" employs pretext tasks derived solely from unsupervised data. These pretext tasks are designed to learn the meaningful visual representation in order to complete them effectively \citep{chen2020simple}. Whereas, in the supervised learning paradigm, each input sample is connected to its matching target or output label, and the system is trained on a labeled dataset. In order for the model to accurately predict outcomes on data that has not yet been observed, supervised learning aims to establish a relationship between input data and their associated output labels \citep{pan2009survey}.

\section{Comparative Analysis of SSL Mechanisms}\label{sec5}





In this section, we will provide a comparative analysis of state-of-the-art SSL (SSL) mechanisms employing ViTs.  Various SSL methods used different networks as their backbone networks. For an unbiased comparison, we compare different SSL methods employing ViT-B/16 \citep{dosovitskiy2020image} as their backbone network.

\subsection{Performance Comparison}
We compare SOTA SSL methods,  in Table \ref{tbl_ssl_fine_tune_linear_prob}, where self-supervised pre-training is performed over ImageNet-1K \citep{ILSVRC15} training dataset. Later, fully supervised training is performed to evaluate the feature representations using (i) linear probing and (ii) end-to-end fine-tuning. The inputs are resized to 224x224 crops and report top-1 validation accuracy.
From Table \ref{tbl_ssl_fine_tune_linear_prob}, we notice that DINO \citep{grill2020bootstrap}, MoCo-V3 \citep{chen2021mocov3}, MSG-MAE \citep{tukra2023improving}, iBOT \citep{zhou2021ibot}, and EsViT \citep{li2021efficient} achieve more than 75\% top-1 accuracy, in terms of fine-tuning using linear probing. It is notable that the top-performing  EsViT \citep{li2021efficient} employed Swin-B \citep{liu2021swin} architecture as the backbone network. In contrast, almost all the methods obtain over 82.0\% in terms of end-to-end fine-tuning. However, MSG-MAE \citep{tukra2023improving}, TEC \citep{gao2022towards}, and dBOT \citep{liu2022exploring} present the top-1 accuracy of 85.3\%, 84.7\%, and 84.5\%, respectively.


\begin{table}[!htbp]
\centering
\caption{Comparison of state-of-the-art methods with prior self-supervised pre-training over ImageNet-1K using 100\% of the labels. The results are reported in terms of top-1 accuracy (\%) using linear probing and fine-tuning. The best three results are in \textbf{\textcolor{red}{red}}, \textbf{\textcolor{green}{green}}, and \textbf{\textcolor{blue}{blue}}.}
\begin{tabular}{|l|c|c|c|c|}
\hline
Methods  & Architecture & Epoch & Linear Probing & Fine-Tuning \\ \hline
SwAV \citep{caron2020unsupervised} & ViT-B/16 & 300 & 71.6 & - \\
SimCLR \citep{chen2020simple} & ViT-B/16 & 300 & 73.9 & - \\
BYOL \citep{grill2020bootstrap} & ViT-B/16 & 300 & 73.9 & - \\
DINO \citep{caron2021emerging} & ViT-B/16 & 300 & \textbf{\textcolor{blue}{78.2}} & 82.8 \\
Data2Vec \citep{baevski2022data2vec} & ViT-B/16 & 800 & - & 84.2 \\
MoCo-v3 \citep{chen2021mocov3} & ViT-B/16 & 300 & 76.7 & 83.2 \\
BEiT \citep{bao2106beit} & ViT-B/16 & 800 & 56.7 & 83.2 \\
MAE \citep{he2022masked} & ViT-B/16 & 800 & 64.4 & 83.4 \\
BootMAE \citep{dong2022bootstrapped} & ViT-B/16 & 300 & 64.1 & 84.0 \\
MSG-MAE \citep{tukra2023improving} & ViT-B/16 & 300 & 75.6 & \textbf{\textcolor{red}{85.3}} \\
TEC \citep{gao2022towards} & ViT-B/16 & 300 & - & \textbf{\textcolor{green}{84.7}} \\
Deit-III \citep{touvron2022deit} & ViT-B/16 & 800 & - & 83.8 \\
Deit-III+AugMask \citep{heo2023augmenting} & ViT-B/16 & 800 & - & 84.2 \\
EsViT \citep{li2021efficient} & Swin-B \citep{liu2021swin} & 300 & \textbf{\textcolor{red}{80.4}} & - \\
iBOT \citep{zhou2021ibot} & ViT-B/16 & 1600 & \textbf{\textcolor{green}{79.5}} & 84.0 \\
dBOT \citep{liu2022exploring} & ViT-B/16 & 800 & - & \textbf{\textcolor{blue}{84.5}} \\
CIM-RevDet \citep{fang2022corrupted} & ViT-B/16 & 300 & - & 83.3 \\
MFM \citep{xie2022masked} & ViT-B/16 & 300 & - & 83.1 \\
MixedAE \citep{chen2023mixed} & ViT-B/16 & 300 & - & 83.8 \\
MaPeT \citep{baraldi2023learning} & ViT-B/16 & 300 & 73.5 & 83.6 \\
\hline
\end{tabular}%

\label{tbl_ssl_fine_tune_linear_prob}
\end{table}

\begin{table}[!htbp]
\centering
\caption{Semantic segmentation ($mIoU$) comparison on ADE20K using Upernet and ViT-B networks. In addition, object detection ($ AP^{box}$) and instance segmentation ($AP^{mask}$) over MS COCO datasets using Cascaded Mask R-CNN with ViT-B.  The best three results are in \textbf{\textcolor{red}{red}}, \textbf{\textcolor{green}{green}}, and \textbf{\textcolor{blue}{blue}}.}
\label{tbl:generaizibility_SSL}
\begin{tabular}{|l|c|c|c|}
\hline
Methods          & $mIoU$ & $AP^{box}$ & $AP^{mask}$ \\ \hline
MoCo             & 47.3                         & 44.9                     & 40.4                           \\
BEiT     \citep{bao2106beit}          & 47.1                         & 49.8                     & 44.4                           \\
DINO   \citep{caron2021emerging}          & 44.1                         & 50.1                     & 43.4                           \\
MoCo-v3      \citep{chen2021mocov3}    & 47.3                         & 47.9                     & 42.7                           \\
MAE   \citep{he2022masked}           & 47.6                         & 46.8                     & 41.9                           \\
BootMAE  \citep{dong2022bootstrapped}        & 49.1                         & 48.5                     & 43.4                           \\
Diet-III  \citep{touvron2022deit}       & \textbf{\textcolor{blue}{49.7}}                         & 50.7                     & 43.6                           \\
Diet-III+AugMask \citep{heo2023augmenting} & \textbf{\textcolor{red}{50.2}}                         & 50.9                     & 43.9                           \\
MAE   \citep{he2022masked}           & 48.1                         & 50.3                     & 44.9                           \\
MSG-MAE   \citep{tukra2023improving}       & \textbf{\textcolor{blue}{49.7}}                         & \textbf{\textcolor{blue}{52.3}}                     & \textbf{\textcolor{red}{48.8}}                           \\
data2vec \citep{baevski2022data2vec}        & 48.2                         & -                        & -                              \\
TEC  \citep{gao2022towards}            & \textbf{\textcolor{green}{49.9}}                         & \textbf{\textcolor{red}{54.6}}                     & \textbf{\textcolor{green}{47.2}}                           \\
iBOT   \citep{zhou2021ibot}          & 48.4                         & 51.2                     & 44.2                           \\
dBOT    \citep{liu2022exploring}         & 49.5                         & \textbf{\textcolor{green}{52.7}}                   & \textbf{\textcolor{blue}{45.7}} \\
MixedAE   \citep{chen2023mixed}   & 48.9      & 51.0                     & 44.1 \\
CIM-RevDet \citep{fang2022corrupted}   & 43.5                         & -                     & - \\
MFM \citep{xie2022masked}  & 48.6                         & -                     & - \\
\hline   
\end{tabular}

\end{table}

\subsection{Performance evaluation over downstream tasks}
We further assess the effectiveness of self-supervised methods for downstream tasks such as object detection, semantic segmentation, and instance segmentation in Table \ref{tbl:generaizibility_SSL}. We also  demonstrate the generalizability capabilities of visual representations learned by self-supervised  methods  by fine-tuning the
pre-trained models on smaller datasets  in Table \ref{tbl_transferability_ssl}.

\subsubsection{Object detection and instance segmentation}
We present a comparative analysis of SOTA, a self-supervised approaches for downstream tasks such as object detection and instance segmentation (as shown in Table \ref{tbl:generaizibility_SSL}. For an unbiased evaluation, we compare techniques having ViT-B \citep{dosovitskiy2020image} as the backbone architecture. The methods are trained over COCO \citep{lin2014coco} dataset considering Mask R-CNN \citep{cai2019cascade} as the task head for object detection and instance segmentation and evaluated using average precision $AP^{box}$ and $AP^{mask}$, respectively. The results in Table \ref{tbl:generaizibility_SSL} demonstrate that TEC \citep{gao2022towards}, dBOT    \citep{liu2022exploring},  and   MSG-MAE   \citep{tukra2023improving} outperform other SOTA SSL methods over object detection and obtain $AP^{box}$ scores of 52.3, 52.7, and 52.3 respectively. On the instance segmentation, MSG-MAE   \citep{tukra2023improving}, TEC \citep{gao2022towards}, and  dBOT    \citep{liu2022exploring} show significant promising performance compared to other SSL methods and achieve 48.8, 47.2, and 45.7  $AP^{mask}$ scores, respectively.

\subsubsection{Semantic Segmentation Comparison}
In Table \ref{tbl:generaizibility_SSL}, we present a performance comparison of SSL methods for semantic segmentation,  where UpperNet \citep{xiao2018unified} is trained with ViT-B \citep{dosovitskiy2020image}  over ADE20K \citep{zhou2017scene} dataset.
From Table \ref{tbl:generaizibility_SSL}, it is notable that BootMAE \citep{dong2022bootstrapped}, Diet-III \citep{touvron2022deit}, Diet-III+AugMask \citep{heo2023augmenting}, MSG-MAE \citep{tukra2023improving}, TEC   \citep{gao2022towards} and dBOT \citep{liu2022exploring} exhibit above 49.0\% mIoU. We also observe that top performing Diet-III+AugMask \citep{heo2023augmenting},  TEC   \citep{gao2022towards}, Diet-III \citep{touvron2022deit}, and  MSG-MAE \citep{tukra2023improving},  demonstate the 50.2\%, 49.9\%,  49.7\%, and 49.7\%, respectively.

\begin{table}[!htbp]
\centering
\caption{Comparison of transfer classification accuracy over CIFAR-10, CIFAR-100, iNat18, iNat19, Flowers, and Cars using ViT-B as backbone network. The best results are in bold.}
\begin{tabular}{|l|c|c|c|c|c|c|c|c|c|} \hline
Methods  & Architecture & CIFAR-10 & CIFAR-100 & iNat18 & iNat19 & Flowers & Cars   \\ \hline 
DINO \citep{caron2021emerging}     & ViT-B/16     & 99.1    & 91.7     & 72.6   & 78.6   & 98.8    & 93.0   \\
MSN \citep{assran2022masked}      & ViT-B/16     & 99.0    & 90.5     & 72.1   & 78.1   &   -   & - \\
iBOT \citep{zhou2021ibot}     & ViT-B/16     & 99.2    & 92.2     & 74.6   & 79.6   & \textbf{98.9}    & \textbf{94.3} \\
MAE \citep{he2022masked}     & ViT-B/16     &   -   &   -   & 75.4   & 80.5   &   -  & - \\
dBOT  \citep{liu2022exploring}   & ViT-B/16     & \textbf{99.3}    & 91.3     & \textbf{77.9}   & \textbf{81.0}   & 98.2    & 93.7 \\
MoCo-v3  \citep{chen2021mocov3}  & ViT-B/16     & 98.9    & 90.5     &  -   &        & 97.7     & - \\
MixedAE \citep{chen2023mixed} & ViT-B/16     & 97.9    & 85.9     &    -  & -    & 97.1     &  88.8\\
Diet-III \citep{touvron2022deit} & ViT-B/16     & \textbf{99.3}    & \textbf{92.5}     & 73.6   & 78.0   & 98.6    & 93.4 \\ \hline
\end{tabular}
\label{tbl_transferability_ssl}
\end{table}

\subsubsection{Transfer Classification Comparison}
To further study the generalizability of the SSL methods, we compare their transfer learning capabilities in terms of accuracy. To do so,  the pre-trained methods are fine-tuned over small datasets such as CIFAR-10  \citep{krizhevsky2009cifar}, CIFAR-100  \citep{krizhevsky2009cifar}, iNaturalist18 \citep{van2018inaturalist} (iNa18), iNaturalist19  \citep{van2018inaturalist} (iNa19), Flowers \citep{nilsback2008automated}, and Cars \citep{krause20133d}. The results in Table \ref{tbl_transferability_ssl} Deit-III \citep{touvron2022deit} demonstrate outstanding performance over state-of-the-art SSL methods and achieve 99.3 and 92.5 accuracy over CIFAR-10 and CIFAR-100 datasets, accordingly. We also note that dBOT  \citep{liu2022exploring}  illustrates better performance over 
CIFAR-10, iNa18, and iNa19 datasets, and show accuracy of 99.3, 77.9, and 81.0, respectively. Similarly, iBOT \citep{zhou2021ibot} exhibits the best performance over Flowers and Cars datasets while obtaining 98.9 and 94.3 accuracy, respectively. 

\subsection{Robustness to Perturbations}
We also present a comparative comparison of SSL methods over various robustness datasets including natural adversarial examples \citep{hendrycks2021natural} (IN-A), objects in different styles and textures (IN-R \citep{hendrycks2021many}), controls in rotation, background, and viewpoints (ObjNet \citep{barbu2019objectnet}), and SI-scores \citep{djolonga2021robustness} (SI-size, SI-loc, and SI-rot).
The results in Table \ref{tbl_robustness_ssl} show that  Diet-III+AugMask \citep{heo2023augmenting} has better robustness across various robustness metrics.

\begin{table}[!htbp]
\caption{Robustness comparison over various robustness benchmarks using ViT-B as a backbone network. It is noticeable that Diet-III+AugMask \citep{heo2023augmenting} demonstrates better robustness performance. The  results in bold indicate better performance.
}

\begin{tabular}{|l@{}|l@{}|l@{}|l@{}|l@{}|l@{}|l@{}|l@{}|l@{}|l@{}|}
\hline
\textbf{Methods}          & \textbf{IN-1K} & \textbf{IN-V2} & \textbf{IN-Real} & \textbf{IN-A} & \textbf{IN-R} & \textbf{ObjNet} & \textbf{SI-size} & \textbf{SI-loc} & \textbf{SI-rot} \\ \hline 
Diet-III    \citep{touvron2022deit}     & 83.8  & 73.4  & 88.2    & 36.8 & 54.1 & 35.7   & 58.0    & 42.7   & 41.5   \\
Diet-III+AugMask \citep{heo2023augmenting} & \textbf{84.2}  & \textbf{74.0}  & \textbf{88.6}    & \textbf{41.9} & \textbf{54.4} & \textbf{37.2}   & \textbf{59.0}    & \textbf{44.8}   & \textbf{43.3}   \\
MFM     \citep{xie2022masked}         &       &       &         & 32.7 & 48.6 &        &         &        &        \\
MAE       \citep{he2022masked}        &       &       &         & 31.5 & 48.3 &        &         &        &     \\ 
\hline
\end{tabular}
\label{tbl_robustness_ssl}
\end{table}





\section{Advancements and Open Challenges}\label{sec6}




In this section, we will discuss the recent advancements and open challenges in SSL (SSL) for ViTs.

\subsection{Data Augmentation as supportive technique }

Data augmentation is a crucial aspect of SSL for ViTs. Data augmentation strategies can assist the SSL models to achieve more robustness and generalizability representations from unlabeled data. Recent advancements in the field including generative models to generate augmented data, incorporating adversarial perturbations, and unsupervised clustering to generate diverse augmentations, to enhance learning capability of the model. However, there is still a need for more effective and productive data augmentation techniques which can help to improve the effectiveness of SSL methods.
In supervised and SSL environments in ViTs, data scarcity is one of the challenges for convolutional neural networks (CNN) models training and generalization. Usually, various conventional strategies of data augmentation such as rotation, contrast enhancement, cropping, and flipping are used in literature for improvement of the learning and generalization ability of the models.   
The SSL approaches learned contextual representation to predict labels without any human interferences and learned the transformation from input data. In SSL the self-supervised labels are optimized by employing two loss functions namely, the original and the self-supervised, that utilized the feature space sharing strategy.  There are mainly two types of data augmentation strategies: (i) traditional and (ii) advanced. Detail of these techniques are given as follows:
\subsubsection{Traditional Augmentation}
Augmentation using geometric transformation is being used for various image processing applications.  By employing this transformation, the original positions of the image pixels are relocated to new positions without modifying intensity values. It modifies the training data by incorporating variations in viewpoint, non-rigid deformations, and scaling for better model learning on real world problems. The geometric transformations including the translation, rotation, reflection, shearing, and scaling are being commonly used for data augmentation. It is categorized into affine and non-affine transformations. The former one utilized linear mapping for adjusting the geometric disorder in the structure of the image. Whereas the later involved complex non-linear mapping functions \citep{gallier2011geometric,struik2011lectures}.
The projective and perspective transformations are fall into the non-affine geometric transformations. These are effectively being used in computer vision and medical imaging tasks \citep{wang2019perspective,franke2021bounding}. 
Similarly, the deformation transformations are non-affine which are used to simulate at higher degree of freedom such as Lens and non-rigid body deformations \citep{milletari2016v,simard2003best,ronneberger2015u}. 
Photometric transformation \citep{xu2023comprehensive, mumuni2022data} is also fall under the traditional augmentation category  and used for various computer vision tasks. It considers the camera and shooting artifacts such as optical noise, motion blur, image color artifacts degradations. These simulations are being used to mitigate the effects of data scarcity during model training.
\subsubsection{Advanced Augmentation Strategies}
The advanced augmentation strategies are used to solve complex computer vision problems and gained a significant attention of researchers. These augmentation operations are learned from the given data automatically. Several CNN based models are proposed in literature to perform various advanced transformations for data augmentation \citep{jaderberg2015spatial,karargyris2015color,tarasiuk2016geometric}. These deep models have several cascading layers to learn the transformations, consequently obtained a better trained and generalized model. Jagerberg et al. (2015) \citep{jaderberg2015spatial} proposed spatial transformer network (STN) data augmentation approach which is used for model training. Based on STN data augmentation model, several improved models are  built which offer promising results \citep{mounsaveng2021learning,luo2020stnreid,vu2017multi}. Although the advanced models are computationally expensive compared to the traditional methods. However, the accuracy and diversity of data augmentation of these models are very high. They are effective to solve various data scarcity related computer vision and SSL problems.
\subsubsection{Challenges of Data Augmentation}
Data augmentation is an imperative task for improvement in performance of the SSL and ViTs. It is successfully applied to solve different computer vision problems. However, there are several open challenges related to uses of data augmentation in SSL and ViTs:
\begin{itemize}
    \item{\emph{Designing of augmentation strategies:} Generally, traditional methods for data augmentation techniques includes flipping, random cropping, and color jittering. It is not a simple task to design a strategy for effective data augmentation in SSL. Exploring novel and appropriate augmentations which exploit certain characteristics of ViTs is an open challenge \citep{cubuk2020randaugment}.}
    \item{\emph{Robustness to complex variations:} Usually, real-world images having complex variations such as in lighting, occlusion, viewpoint, etc. It is a major challenge to validate that during SSL and data augmentation process such complex variations are truly accounted \citep{geirhos2018imagenet}.}
    \item{\emph{Maintaining of consistency and discriminability:} Typically, SSL techniques use produced augmented visions of the same image and capture useful patterns to support consistency that is small changes in the input will reflect in the learned representations. In order to maintain discriminative information, it is challenging to discover the sufficient balance between produced augmented visions consistent for learning \citep{tian2020makes}.}
    \item{\emph{Scale Sensitivity:} ViTs work excellent on large-scale datasets. The data augmentation methods well-behaved on data scarcity problems and it might not be effective on larger-scale datasets. Development of augmentation methods, which are capable to apply on larger-scale datasets is a challenge \citep{dosovitskiy2020image}.}
    \item{\emph{Computational Efficiency:} Commonly, ViTs are computationally intensive, and incorporation of data through augmentation techniques would make problem more intensified. Exploring adequate data augmentation methods which are computationally efficient and offering valuable benefits to SSL is a practical challenge \citep{touvron2019fixing}.}
    \item{\emph{Robust against adversarial attacks:} Data augmentation methods usually are not vulnerable against adversarial attacks. It is open challenge to develop efficient data augmentation approaches which improve developed model robustness against various forms of adversarial attacks \citep{hassan2023developing}.}
\end{itemize}

\subsection{Incorporating Spatial Information}

Incorporating spatial information into SSL-based ViTs is an active area of research nowadays. Spatial details are essential for grasping the structural relationships within images and can improve the model's ability to capture meaningful features. One recent trend incorporating spatial information is utilizing positional embeddings using sine/cosine and axial position embeddings \citep{chu2021conditional}. Spatial relationships can also be better captured using grid masking \citep{li2021mst}. The attention mechanisms can also be used to capture the spatial dependencies, e.g., axial attention \citep{ZHENG2023105587} which can help the model to focus on local and global spatial features. However, incorporating spatial information might be challenging due to increase the computational complexity of transformers. Further, it complicates when dealing with multi-modal data or tasks that require understanding spatial-temporal relationships. Therefore, spatial attention is one of the components of SSL and ViTs that can be encountered  when solving a complex problem.

\subsection{Improving Sample Efficiency}
The performance of sample efficiency can be improved by following some of the good practices. For instance, bigger training datasets can be further divided into subsets to capture fine details of input data. If the training dataset is already small, then labeled data can be combined with unlabeled data to perform SSL learning. Regularization methods such as dropout, weight decay, and layer normalization can also improve sample efficiency. Regularization helps to prevent overfitting and improve the model's generalization. It is recommended to employ the transfer learning concept from larger, pre-trained models for training of smaller ViTs to mimic the behaviour of the larger models \citep{wang2022attention}. By following this practice, smaller model's performance can be improved while having limited dataset. Fine-tuning strategies such as layer freezing, gradual unfreezing, and learning rate schedules can also affect the sample efficiency.

\subsection{Interpretability and Explainability}

Interpretability and explainability are critical for the utilization and veracity of the models.  To understand the inner working mechanism of the model while solving real world problems, the following steps might be considered while designing the ViTs models. For interpretability, the important aspects include feature visualization, activation analysis, and attribution methods can be considered during model development \citep{zeiler2014visualizing,zhu2017unpaired,selvaraju2017grad} . While, for explainability of ViTs models, the attention mechanism, attention maps, feature attribution, concept activation, and shapely values need to be accounted \citep{koh2017understanding,guidotti2018survey,lundberg2017unified}.

In summary, recent advancements and open challenges in SSL for ViTs include improving data augmentation strategies, incorporating spatial information, improving sample efficiency, and improving interpretability and explainability. Addressing these challenges will enable the development of more effective and efficient SSL methods for ViTs.

\section{Future Directions}\label{sec7}




The exploration of various innovative concepts in VIT architectures has transformed the course of research, particularly in the fields of image processing and computer vision.
\begin{itemize}
\item Current SSL techniques in vision often rely on specific pretext tasks (e.g., rotation prediction, contrastive learning), which may not generalize well in few-shot settings. Future work could focus on designing task-agnostic or adaptive SSL pretext tasks that retain performance even with limited data and across domains.

\item While some methods use SSL before meta-learning, a tighter integration between SSL and meta-learning algorithms (rather than sequential use) might unlock stronger generalization in low-data regimes. For instance, end-to-end joint optimization could be explored.

\item Cross-domain FSL remains a significant challenge due to large distributional differences between base and novel classes. Research should explore domain-invariant SSL strategies and transferable feature spaces that maintain discriminative power across domains.

\item While transductive approaches like MoCo v2 have shown promise, they often rely on access to test data, which violates the true few-shot learning setting. Future methods could aim to simulate transductive benefits without requiring test-time class exposure e.g., via generative models or domain-agnostic augmentation strategies.

\item Unsupervised FSL methods (like clustering + meta-learning) underperform compared to supervised ones. There's potential in developing stronger pseudo-label generation, iterative self-labeling, or semi-supervised hybrid pipelines that bridge the gap between unsupervised and supervised methods.

\item The InfoMax principle is promising, but current mutual information estimators may introduce bias. Future work should explore more accurate or task-adaptive MI estimation techniques to better capture dependencies in few-shot setups.

\item A comprehensive benchmark that includes varied domains, pretext tasks, and evaluation protocols would allow for more meaningful comparisons of SSL-enhanced FSL methods and help identify what truly drives performance.

\item SSL often relies on the data distribution of the unlabeled dataset. However, zero-shot learning requires the model to generalize to unseen classes, which may not follow the same distribution. In real-world applications, the semantic attributes used for zero-shot learning may contain noise or inaccuracies, affecting the performance of the combined model. By addressing these challenges, the combined approach of ZSL and SSL can pave the way for more generalize and efficient vision models, thereby pushing the limits of what can be accomplished in the field of computer vision.

\item Real-world deployment is a crucial aspect of SSL for ViTs. SSL methods should be efficient, scalable, and effective in real-world scenarios. Future directions in real-world deployment include exploring the effectiveness of SSL methods on more diverse and complex datasets, developing more efficient and scalable SSL methods, and addressing the ethical and societal implications of SSL methods in real-world applications.
\end{itemize}

\begin{figure}[t]
      \centering
      \includegraphics[width=10cm]{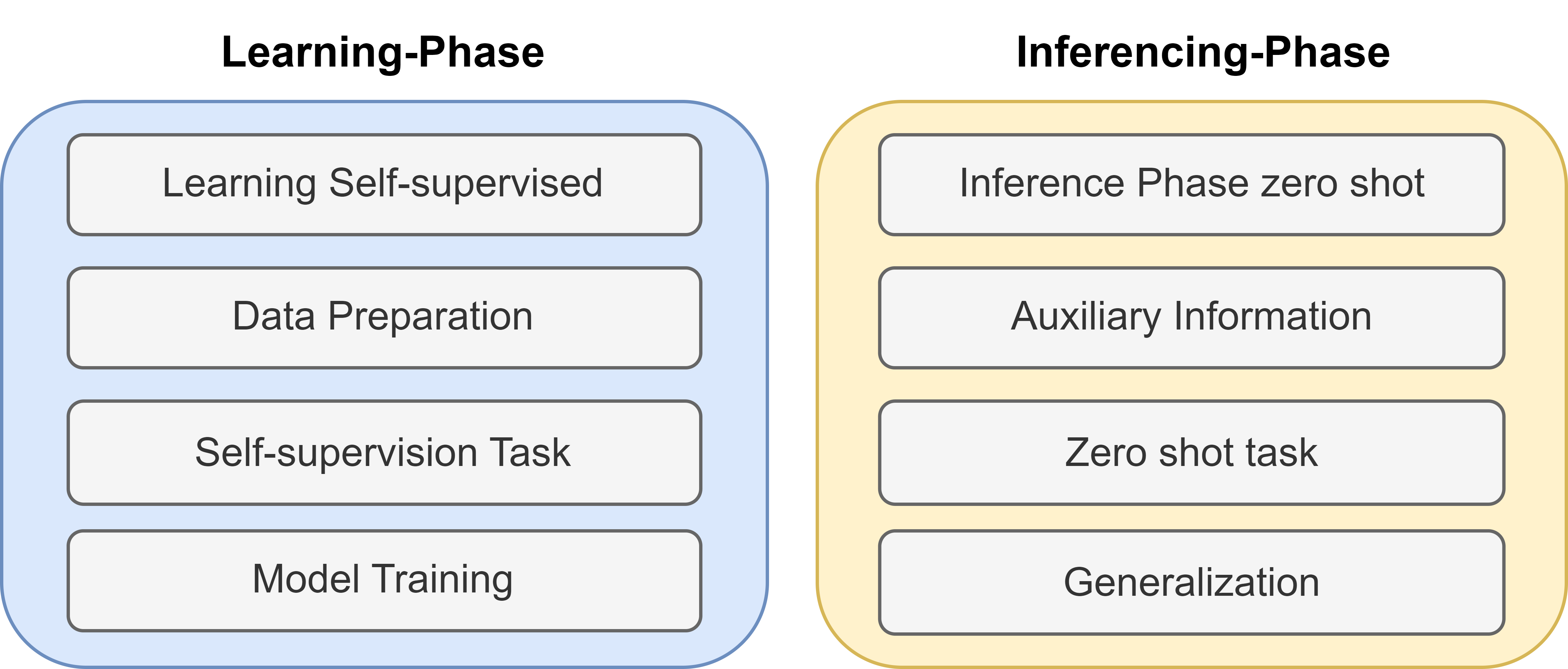}
      \caption{The block diagram illustrating the Learning Phase (Self-Supervised) and the Inference Phase (Zero-Shot)}
      \label{ZSL}
\end{figure}
\par

\section{Conclusion}\label{sec13}

This survey paper has provided a comprehensive overview of SSL methods specifically for ViTs. The taxonomy of SSL methods has been illustrated, highlighting the four major categories as contrastive, generative, predictive, and hybrid methods. The exploration of SSL extends to a detailed summary of various pretext tasks, including spatial context-based methods, color and texture based methods, temporal and sequence based methods, contrastive and clustering methods, distillation and momentum-based methods, redundancy reduction methods, and cross-modal methods. By delving into these diverse aspects, this survey contributes to a deeper understanding of the landscape of SSL methods, providing valuable insights for both researchers and professionals in the domain of computer vision.

\section*{Statements and Declarations}

\subsection*{Funding}
This research did not receive external funding.

\subsection*{Competing Interests}
The authors declare no conflict of interest.


\bibliography{sn-article}

\end{document}